\documentclass{IEEEtaes}
\usepackage{color,array,amsthm}
\usepackage{graphicx}
\usepackage{amssymb}
\usepackage{amsmath}
\usepackage{multirow}
\usepackage{hyperref}

\newcommand {\R}{\mathbb{R}}

\jvol{XX}
\jnum{XX}
\jmonth{XXXXX}
\paper{1234567}
\pubyear{2020}
\doiinfo{TAES.2020.Doi Number}

\begin{document}

\title{Test-Time Adaptation for Keypoint-Based Spacecraft Pose Estimation Based on Predicted-View Synthesis}

\author{Juan Ignacio Bravo Pérez-Villar}
\affil{Deimos Space, Madrid, Spain. Video Processing and Understanding Lab, Universidad Autónoma de Madrid, Spain} 

\author{Álvaro García-Martín}
\affil{Video Processing and Understanding Lab, Universidad Autónoma de Madrid, Spain} 

\author{Jesús Bescós}
\affil{Video Processing and Understanding Lab, Universidad Autónoma de Madrid, Spain} 

\author{Juan C. SanMiguel}
\affil{Video Processing and Understanding Lab, Universidad Autónoma de Madrid, Spain} 


\receiveddate{}

\corresp{}

\authoraddress{}

\editor{}
\supplementary{}

\markboth{Bravo Pérez-Villar et al.}{TTA for Keypoint-Based Spacecraft Pose Estimation Based on Predicted-View Synthesis}
\maketitle

\begin{abstract}Due to the difficulty of replicating the real conditions during training, supervised algorithms for spacecraft pose estimation experience a drop in performance when trained on synthetic data and applied to real operational data. To address this issue, we propose a test-time adaptation approach that leverages the temporal redundancy between images acquired during close proximity operations. Our approach involves extracting features from sequential spacecraft images, estimating their poses, and then using this information to synthesise a reconstructed view. We establish a self-supervised learning objective by comparing the synthesised view with the actual one. During training, we supervise both pose estimation and image synthesis, while at test-time, we optimise the self-supervised objective. Additionally, we introduce a regularisation loss to prevent solutions that are not consistent with the keypoint structure of the spacecraft. Our code is available at: \href{https://github.com/JotaBravo/spacecraft-tta}{https://github.com/JotaBravo/spacecraft-tta}.

\end{abstract}



\section{Introduction}\label{seq:introduction}

Fully recreating the image conditions encountered in spacecraft proximity operations, either via simulated or hardware-in-the-loop settings, is often unfeasible, presenting uncertainties on the test-time behaviour of the spacecraft pose estimation models. These inherent differences between the train (source) data and the test (target) data distributions, known as \textit{domain-gap}, cause that spacecraft pose estimation methods trained over synthetic computer-generated datasets, experience a drop in performance when evaluated on the operational target data. The introduction of datasets containing both computer-generated and hardware-in-the-loop posed images~\cite{park2022speedplus,cassinis2023leveraging}, have led to numerous improvements to mitigate the effects of the domain gap~\cite{park2023satellite}. A common particularity of such methods is that they rely on single images of the spacecraft. However, during real close-proximity operations pose estimation models process sequential images or views of the target spacecraft with high temporal redundancy. 

\begin{figure}[h]
    \centering
    \includegraphics[width=1\columnwidth]{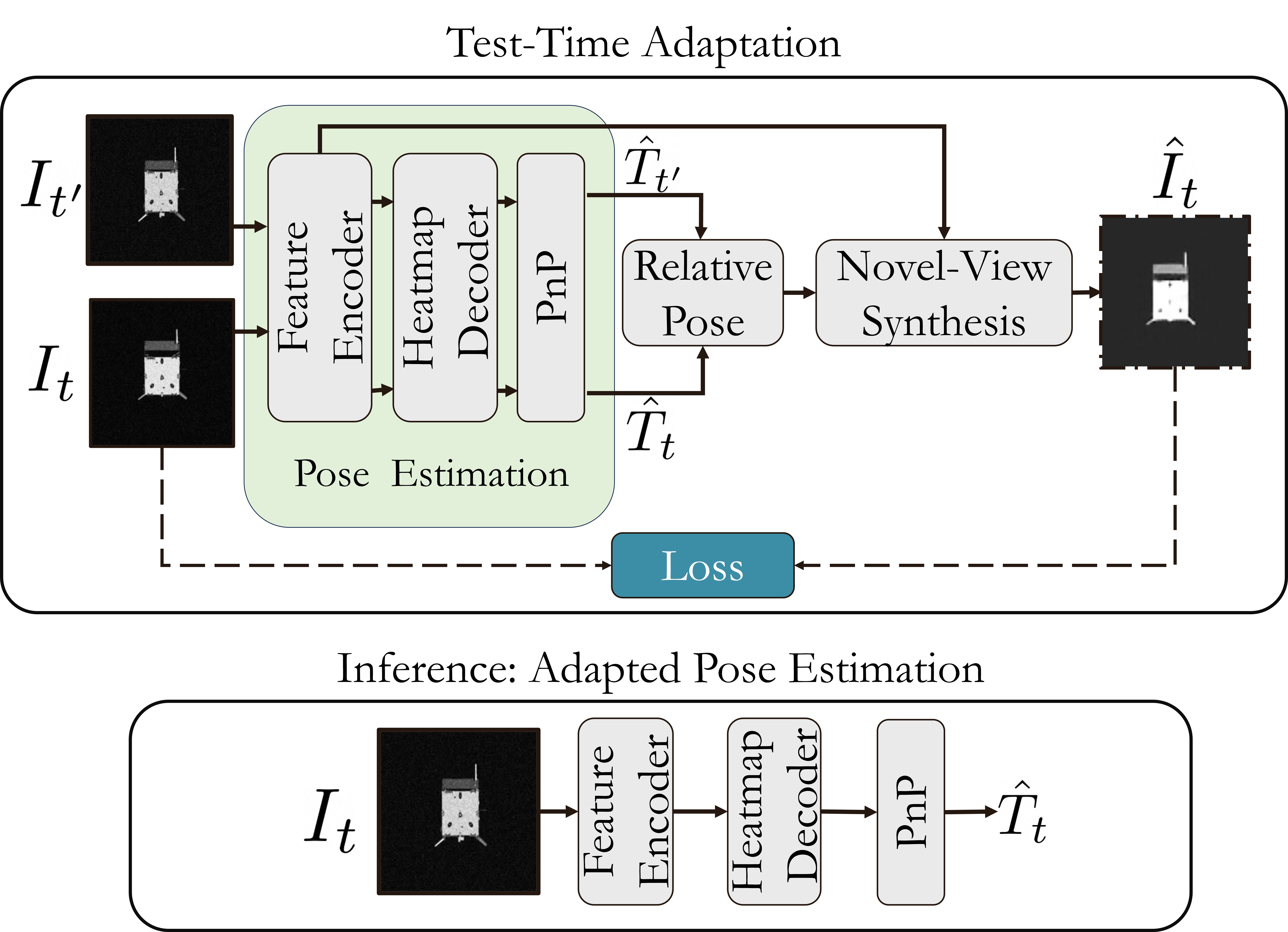}
    \caption{Schematic of the proposed approach for test-time adaptation in spacecraft pose estimation. The process involves retrieving feature representations for two consecutive views of a spacecraft $I_{t'}, I_t$ and estimating their absolute poses. The spacecraft pose variation between the two views is then estimated from these absolute poses. Then, using the feature representation of $I_{t'}$ and the pose variation, $I_t$ image is reconstructed $\hat{I}_t$. During the test-time adaptation stage the photometric difference between the $I_t$  and $\hat{I}_t$ is minimised, in addition to a heatmap regularisation loss. After the test-time-adaptation stage, only the adapted keypoint pose estimation pipeline is employed to obtain the pose of the target spacecraft. During training the model is optimised with the supervised (orange boxes) and self-supervised (blue boxes) learning objectives. At test-time only the self-supervised objective is optimised.}
    \label{fig:introduction}
\end{figure}

We propose to exploit this redundancy to perform test-time adaptation, with the objective of improving the pose estimation performance on new test domains. This test-time adaptation stage is performed via predicted-view synthesis, more generally known as novel-view synthesis~\cite{rematas2016novel}. The core idea of our method is represented in Figure~\ref{fig:introduction}: First, we retrieve a feature representation for two time adjacent views of a spacecraft, and use each feature representation to estimate their respective absolute pose using a keypoint based method. The keypoint-based method focuses on estimating heatmaps that encode the probability of a 3D spacecraft keypoint being located at specific 2D image coordinates. The method then employs Perspective-n-Point (PnP) solver to derive the pose from these 2D-3D correspondences.

Next, from these absolute poses we estimate the relative spacecraft pose variation between the two time instants $t'$ and $t $. Using the feature representation from one view $I_{t'}$ and considering the computed pose variation, we estimate a reconstruction of a second view or image $\hat{I}_{t}$. Since this second image $I_{t}$ is known, we can derive a self-supervised learning objective by measuring the photometric difference between the synthesised view and the actual view. During training we supervise both pose estimation (supervised) and novel-view synthesis networks (self-supervised) and at test-time we optimise the self-supervised objective. In this way we achieve a self-supervised task that is coupled with the pose estimation one. Performing test-time-adaptation with a self-supervised task that is coupled or similar to the supervised task, provides increased performance with those that are not coupled or related~\cite{liu2021ttt++}. This motivates the introduction of our pose-related self-supervised task for test-time adaptation. Finally, to avoid degenerate solutions in the keypoint estimation process that explain the best relative pose between images but are not consistent with the spacecraft structure, we propose a regularisation loss to incorporate the keypoint structure information. After the test-time-adaptation stage, only the adapted keypoint pose estimation pipeline is employed to obtain the pose of the target spacecraft. Our contributions are twofold: 
\begin{itemize}
    \item A self-supervised test-time adaptation framework for spacecraft pose estimation based on novel-view synthesis that explicitly relates the pose estimation with the image reconstruction process.
    \item A heatmap regularisation loss for self-supervised keypoint learning. This loss is designed to avoid degenerate solutions in the keypoint estimation process that explain the best relative pose between images but are not consistent with the spacecraft structure.
\end{itemize}

The rest of the article is organised as follows: In Section~\ref{seq:background} we conduct a review of spacecraft pose estimation methods, we review the existing test-time adaptation approaches for spacecraft pose estimation, and introduce the theoretical background of novel-view synthesis from pose. Next, we describe the details of our method in Section~\ref{seq:proposed-approach}. We run extensive experiments and comparison against other methods in Section~\ref{seq:experiments}, and conclude in Section~\ref{seq:conclusions}.


\section{Background}\label{seq:background}

\subsection{Monocular Spacecraft Pose Estimation}

Monocular spacecraft pose estimation involves determining the rotation $R$ and translation $v$ of a target spacecraft with respect to a chaser spacecraft from a single monocular image. Existing monocular methods can be divided into direct or based on keypoints. For a review of spacecraft pose estimation methods encompassing other electro-optical sensors we refer the interested reader to~\cite{opromolla2017review}.

\textbf{Direct methods} transform two-dimensional images or their feature representations into a six-dimensional output space without the use of geometric solvers. These methods can either discretise the pose space or regress the pose as a continuous output. In~\cite{shi2017spacecraft}, a discrete approach relying on designed features is proposed. This approach involves creating a collection of image representations, each linked to specific pose labels, forming a codebook. This codebook is generated using Principal Component Analysis (PCA). During testing, the PCA representation of an image is calculated and compared against the codebook. The pose of the image is determined by matching the codebook entry that has the closest PCA representation. PCA has also been used in~\cite{zhang2023uav}, where in combination with a Support Vector Machine (SVM) are employed in a trajectory prediction pipeline. A CNN-based discrete method is proposed in~{\cite{sharma2018pose}} where AlexNet is employed to classify input images into discrete poses. Discrete methods offer a straightforward approach to pose estimation, but their effectiveness is limited by the discretization error in the pose space. To address this issue, some works view the output pose space as continuous and approach the problem by regressing a pose directly from an input image. The work of~\cite{proencca2020deep} illustrate this concept by using a CNN to regress the position and estimate the orientation of a spacecraft continuously, employing soft assignment coding. However these methods may struggle to generalise to new domains, due to their formulation often resembles an image retrieval task more than a true pose estimation problem~\cite{sattler2019understanding}.

\textbf{Keypoint-based methods} leverage pre-existing information from a 3D model of the target spacecraft. Let $\{P_i\in \R^4,i=1, \dots, N\}$ be a set of 3D real-world spacecraft keypoints, expressed in homogeneous coordinates (i.e., the coordinates of the Projective Geometry~\cite{hartley2003multiple}) respect to the spacecraft reference frame. Given an image $I$ of the target spacecraft, the goal of keypoint-based methods is to estimate the projected pixel coordinates of $P_i$, denoted as $p_i$. For a given pose of the target spacecraft $T = [R|v]$ respect to a camera defined by the intrinsic parameters $K$, $P_i$ and $p_i$ are related by means of the perspective projection equation:

\begin{equation}
       \lambda_i p_i = K[R|t]P_i.\label{eq:perspective}
\end{equation}

With $\lambda_i$ denoting a scale factor that represents the depth of each point~\cite{wang2022revisiting}. Hence, if the set $P_i$ and the camera parameters $K$ are known, and a good estimate of each $p_i$ is retrieved, the pose can be derived by minimising the following expression:

\begin{equation}
        \text{argmin}_{R,t} \sum^N_{i=1} \lVert \lambda_i p_i -  K[R|t]P_i \rVert_2.
\end{equation}\label{eq:pnp}

The above expression is often solved either with off-the-self PnP estimators or with learned models~\cite{legrand2023end},~\cite{huang2023end}. These are categorized into special PnP solvers and general PnP solvers. Special PnP solvers are designed for a fixed number of points, for instance, P3P, that works for a set of three correspondences (see~\cite{persson2018lambda},~\cite{gao2003complete}). General PnP solvers, in contrast, work with any number of points greater than three and can be further divided into iterative and non-iterative approaches. Iterative methods use nonlinear least squares solvers, examples of which include Procustes PnP~\cite{garro2012solving} and SQPnP~\cite{terzakis2020SQPnP}. Non-iterative methods directly solve the problem without relying on iterative optimization, such as EPnP~\cite{lepetit2009epnp} and UPnP~\cite{penate2013exhaustive}. For a more comprehensive review of PnP methods, we refer the interested reader to~\cite{pan2021survey}. The work from~\cite{chen2020end} provides a backpropagatable PnP estimator that can be used to optimise models in an end-to-end fashion.

A popular approach in the literature is to design methods to derive $p_i$ from the image. Keypoint approaches not based on machine learning make use of tailored feature detectors and descriptors to match 2D image points with a set of previously detected 3D points. This approach is proposed in~{\cite{shi2016spacecraft}} where a set of SIFT~{\cite{lowe2004distinctive}} keypoints and descriptors are employed in combination with a PnP solver to retrieve the spacecraft pose. Similarly, other methods employ lines or wireframe structures to extract the spacecraft pose~{\cite{petit2011vision}}~{\cite{petit2012vision}}. The work of~{\cite{sharma2016comparative}} provides a comparison of design-based approaches that rely on points or lines for the task of spacecraft pose estimation.

Amongst keypoint methods based on machine learning the most common approach is to regress a two-dimensional heatmap that indicates the expected location of each projected keypoint. These methods can be combined with an object detection network to first locate the satellite and then regress the 2D keypoints from the heatmap~\cite{chen2019satellite}. Other works instead propose to derive heatmaps at different resolutions and fuse the information to achieve scale invariance~\cite{hu2021wide}. Including depth information is proposed in~\cite{perez2023spacecraft} to incorporate structure and achieve robustness to domain shift. The use of transformers to derive the heatmaps is explored in~\cite{wang2022revisiting} to explicitly introduce structure in the keypoint learning process. The direct regression of the 2D-bounding box surrounding the spacecraft instead of relying on an intermediate heatmap representation is explored in~\cite{sharma2018pose}.

\subsection{Test-Time Adaptation} 

Test-time adaptation is the process of fine-tuning a pre-trained model on a source domain over a small amount of unlabelled data on the target domain, with the goal of improving the performance of the model on this target domain. The source domain refers to the domain of the data used for training the model, for instance, data that is synthetically generated. In contrast, the target domain encompasses the domain of the data where the model is applied, such as the data collected by spacecraft during real operational conditions. These differences in visual characteristics between domains, known as the domain gap, typically diminishes the effectiveness of models trained on one domain when applied to another. Test-time adaptation aims to mitigate this drop in performance by refining the trained model using unlabelled data from the target domain. Test-time adaptation has the characteristic that train and test data are not jointly observed during training, contrary to other situations in domain adaptation where training and test, unlabelled, data are mutually observed. Methods for test-time adaptation in spacecraft pose estimation are a recent topic of research. We divide the methods into pseudo-labelling and multi-objective learning. \textbf{Pseudo-labelling} methods are based on deriving a small set of confident predictions on the target domain obtained by the model pretrained on the source domain; these confident predictions are then used to fine-tune the model to the target domain. This method is explored in~\cite{perez2023spacecraft}, where pose pseudo-labels are obtained exploiting inter-model consensus: pose predictions are extracted at each level of a stack of two hourglass networks and they are then used as training pseudo-labels if the pose retrieval algorithm at both levels converges. Methods based on \textbf{multi-objective learning} train a model to perform two or more tasks, being some of them supervised and the remaining ones self-supervised. During the test-time adaptation stage the self-supervised objective is used to fine-tune the model to the target domain, with the aim that improvements on the self-supervised tasks transfer to the supervised ones. A multi-objective learning concept is explored in~\cite{park2022robust}; this work proposes the SPNv2, a multi-head network used to predict keypoint heatmaps, spacecraft pose, and a segmentation mask of the spacecraft. During the test-time adaptation, the entropy of the predicted segmentation mask is minimised to adapt the normalisation layers of the feature encoder. A combination of pseudo-labelling and multi-objective learning is explored in~\cite{wang2022bridging}, where pseudo-heatmaps and pseudo-segmentation masks are used to perform domain adaptation. Their approach also includes the use of a CycleGAN to convert the source synthetic images to the target domain data, requiring that both datasets are observed during training. 

All the strategies described help reduce the effect of the domain gap relying on single images of the target spacecraft, but do not incorporate knowledge derived from the temporal redundancy between the images, which is present in a real rendezvous scenario. In this work we aim to incorporate such knowledge by entangling the pose estimation with novel-view synthesis.

\subsection{Novel-View Synthesis From Pose}

Given an image of a scene in a specific pose, novel-view synthesis is the task of generating an image of the same scene in a different target pose. The pixel coordinates corresponding to these poses, $p_t$ and $p_{t'}$, captured with a camera $K$, are related by means of the ego-motion of the chaser spacecraft, expressed as the relative camera pose $T_{t\rightarrow t'}$, and the depth of the scene $D_{t}$~\cite{zhou2017unsupervised}. The reprojected pixel coordinates of $p_{t}$ onto $p_t'$ can be obtained by:

\begin{equation}
    p_{t'} =KT_{t\rightarrow t'}D_{t}(p_{t})K^{-1}p_{t}.\label{eq:image-reconstruction}
\end{equation}

Intuitively, Equation~\ref{eq:image-reconstruction} expresses that the pixel coordinates of a frame $t$ can be transformed to those of a target frame $t'$ by first lifting the $p_{t}$ coordinates into the 3D world coordinates, then rotating and translating the resulting point cloud to the $t'$ coordinate system, and finally projecting the point cloud to the target frame coordinates. This expression relates the novel-view synthesis problem with the problem of estimating the pose and the depth between two frames. This coupling of pose and depth to novel view synthesis is exploited in self-supervised monocular depth estimation~\cite{zhou2017unsupervised, godard2019digging} problems, or neural radiance fields~\cite{mildenhall2021nerf}. In this work we exploit this coupling to define a self-supervised objective that relates the pose estimation problem with novel-view synthesis. We employ the novel-view synthesis to derive a self-supervised objective, which is optimised during the test-time adaptation stage. This adaptation aims to improve the accuracy of our spacecraft pose estimation pipeline. More details on this process are provided in Section~\ref{seq:proposed-approach}.


\section{Proposed approach}\label{seq:proposed-approach}

In this section, we introduce our proposed approach in an incremental manner. Initially, we present the pose estimation method, followed by the test-time adaptation stage, and conclude by outlining the overall method. This study focuses on creating a framework for test- time adaptation in spacecraft pose estimation methods using novel-view synthesis. We have adopted reliable and established baselines for pose estimation~\cite{xiao2018simple} to ensure a solid foundation. Our primary goal is to evolve a replicable and robust framework that can serve as a strong basis for future advancements.

\subsection{Pose Estimation}\label{seq:pose-estimation}

\begin{figure}
    \centering
    \includegraphics[width=\columnwidth]{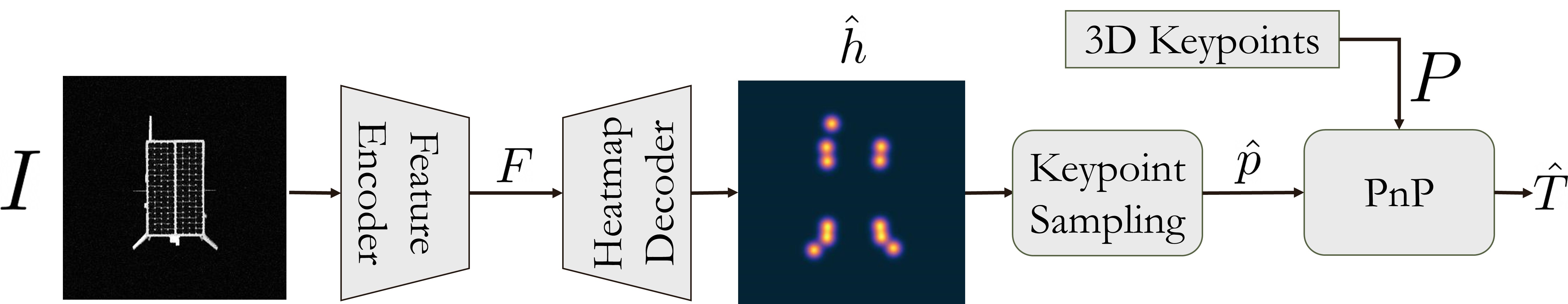}
 \caption{Pose estimation pipeline. An image $I$ is processed with a feature encoder to obtain a feature representation $F$. This feature vector is fed to a heatmap decoder to obtain a heatmap $\hat{h}$ that encodes the probability of the spacecraft 3D keypoints $P_i$ to be in the 2D image coordinates $p$. The estimated keypoint positions $\hat{p}$ are extracted as the coordinates of $\hat{h}$ with maximum probability. The pose estimate $\hat{T}$ is obtained from the 2D-3D correspondences with a PnP solver.}
    \label{fig:pose-estimation-pipeline}
\end{figure}

The schema of the keypoint-based pose estimation architecture is depicted in Figure~\ref{fig:pose-estimation-pipeline}. The goal of the method is to estimate the 2D image coordinates $\hat{p}$ of the predefined 3D object keypoints $P$. The set of 2D-3D correspondences are employed over a PnP solver to retrieve the estimated pose $\hat{T}$. The complete pipeline is as follows: given an image $I$ of the target spacecraft, we first retrieve a feature representation $F$ with a feature encoder. The feature representation of the image $F$ is fed to a heatmap decoder to regress a heatmap $\hat{h} \in R^{H\times W\times N}$, where $N$ represents the channel dimension, each belonging to an unique keypoint, and $W$, $H$ represent the width and height spatial dimensions of the heatmap respectively, that match those of the input image $I$. The channel $i=1,...N$ of the heatmap encodes the probability of a spacecraft 3D keypoint $P_i$ to be in the 2D image coordinates $p_i$. At test time, the estimated image coordinates of a keypoint $\hat{p}_i$ are obtained as the coordinates of the maximum value in the heatmap channel $i$ of $\hat{h}$, we refer to this process as keypoint sampling. With the set of 2D-3D correspondences, the pose estimate $\hat{T}$ is retrieved with a PnP method. 

The feature encoder and heatmap decoder, are supervised with the mean squared error between the ground-truth heatmap $h$ and the estimated heatmap $\hat{h}$

\begin{equation}
    \ell_{h} = \frac{1}{WHN} \Vert \hat{h} - h \rVert_F^2. \label{eq:loss_heatmap}
\end{equation}

Where the pedix $F$ expresses the Frobenius norm. To build the ground-truth heatmaps $h$ for supervision, we first derive the ground-truth keypoint coordinates $p_i$. Given a ground-truth pose $T$ and a camera defined by the intrinsics $K$, the ground-truth coordinates $p_i$ are obtained by projecting the 3D keypoints $P_i$ via Equation~{\ref{eq:perspective}}. Next, the ground-truth heatmap for the channel $i$ is generated by representing a circular 2D Gaussian with fixed standard deviation ($\sigma=7$px) centred at the ground-truth coordinates $p_i$. We choose a value of $\sigma=7$px to ensure that the heatmap coverage area remains consistent with the original work by \cite{tompson2014joint}.

\subsection{Test-Time Adaptation Via Novel-View Synthesis} \label{seq:test-time adaptation-novel-view}

In situations with a large domain shift, keypoint-based approaches often face challenges in retrieving accurate heatmaps, hence resorting to an additional adaptation stage~\cite{park2023satellite}. In this work we perform test-time adaptation by joint optimisation of a supervised and a self-supervised task. During training, both tasks are optimised simultaneously to achieve a shared feature representation. In the presence of domain shift, we update the model via the self-supervised task to align the features with the target test domain, with the ultimate goal of improving the performance of the original supervised task on this new domain.

A self-supervised task for test-time adaptation yields better results the closer it aligns with the supervised task~{\cite{liu2021ttt++}}. Our supervised task is to accurately estimate the heatmap encoding the spacecraft image keypoint coordinates, that are employed to retrieve the spacecraft pose. Thus, we aim to design a self-supervised training task that is related to both keypoint retrieval and pose estimation.

\begin{figure}
    \centering
    \includegraphics[width=1\columnwidth]{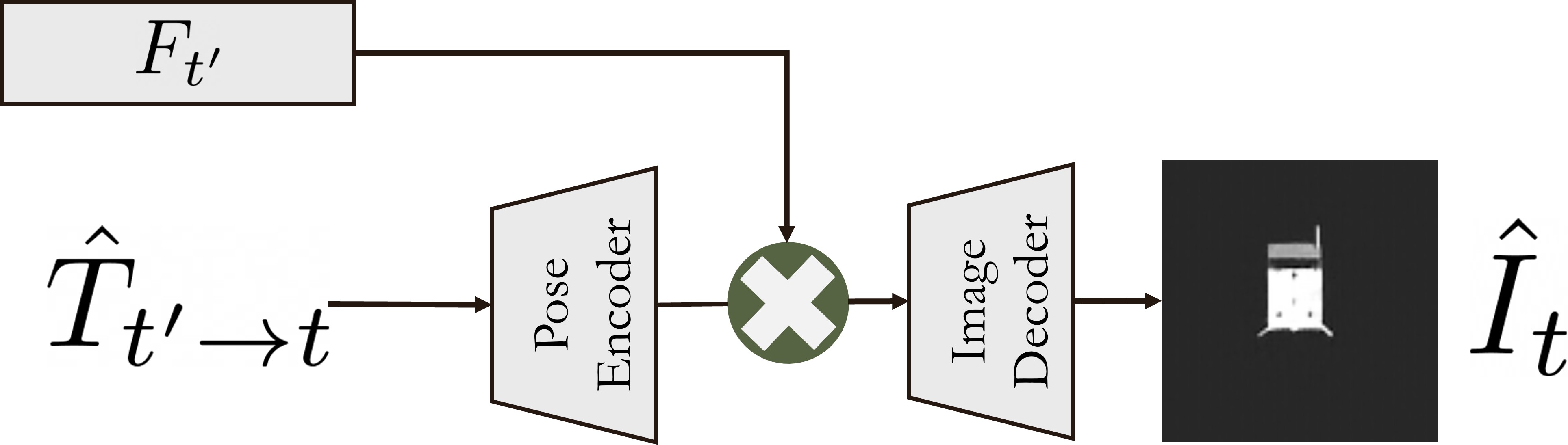}
    \caption{Novel-view synthesis pipeline. The estimated relative pose $\hat{T}_{t' \rightarrow t}$ the spacecraft between two time instants $t'$ and $t$ is fed to an pose encoder and the resulting feature vector is concatenated to the feature vector $F_{t'}$ resulting on encoding the image $I_t'$. The concatenated features are fed to an image decoder to obtain a prediction of the image on the $t$ time instant $\hat{I}_t$. \includegraphics[scale=0.2]{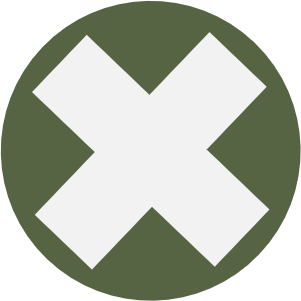} represents concatenation.}
    \label{fig:novel-view-synthesis}
\end{figure}

The task of novel-view synthesis involves creating new images that depict an object from a distinct point of view. As shown in Section~\ref{seq:background}, two views of a scene are related by the relative camera pose and the scene depth (distance of any element of the scene to the camera viewpoint). Learning both depth and pose renders in this case unfeasible as these methods assume that the photometric difference between two images is explained by the camera motion, assumption that is broken by the presence of reflective surfaces in the spacecraft and the relatively low signal-to-noise ratio of the images~\cite{zhou2017unsupervised}. To overcome this limitation, and inspired on the work of~\cite{li2021test}, we choose to train a model that receives a feature representation of an image $F_{t'}$ and the estimated relative camera pose $\hat{T}_{t' \rightarrow t}$ between two images $I_{t'}$ and $I_t$ and estimates a novel view of the scene $\hat{I}_t$ (see Figure~\ref{fig:novel-view-synthesis}). Then, the photometric difference between the estimated image $\hat{I}_t$ and the actual image $I_t$ provides a self-supervised objective related to the pose estimation:

\begin{equation}
    \ell_I = \frac{1}{WH}\sum_{i=1}^{WH}|\hat{I}_t(i) - I_t(i)|.
\end{equation}

Prior to concatenate the camera poses with the image features $F_{t'}$ we encode the camera poses to increase the dimensionality, encouraging that the amount of information contained by the pose and the image representation are balanced.

To improve the pose estimation performance over a new domain, the process begins with estimating the absolute spacecraft pose at two adjacent time instants. Next, the relative pose between these instants is determined. The feature vector of image $I_{t’}$, combined with the estimated relative pose change $ \hat{T}_{{t’} \rightarrow t} $, is used to predict the view $\hat{I}_{t}$. The photometric difference between this predicted view and the actual view serves as a self-supervised objective during test-time training. This procedure uses the same feature vector for pose estimation (achieved through heatmap estimation) and the self-supervised objective. Enhancing the self-supervised objective leads to better alignment of features with the new test domain, which in turn improves the outcomes of the supervised pose estimation objective.

\subsubsection{Heatmap Regularisation}

\begin{figure}
    \centering
    \includegraphics[width=\columnwidth]{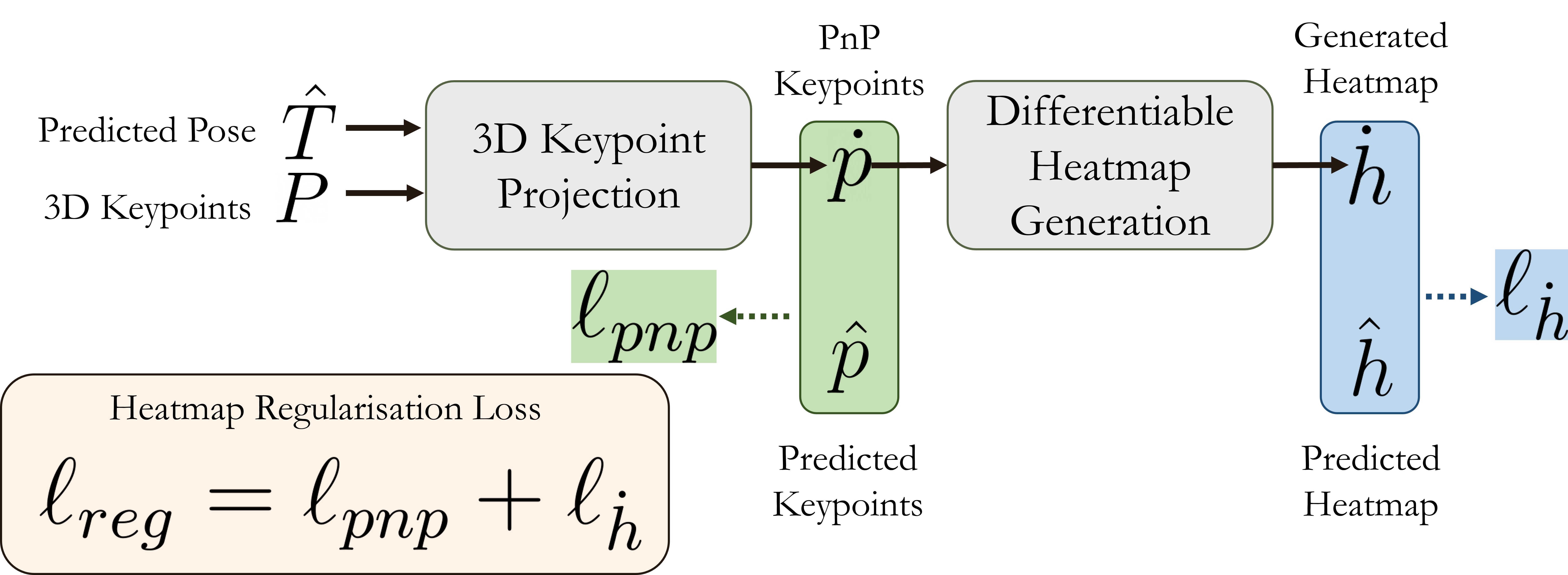}
    \caption{Proposed heatmap regularisation losses: First, the predicted pose using PnP $\hat{T}$ is utilised to project the 3D spacecraft keypoints $P$, yielding the PnP-estimated 2D keypoint coordinates $\dot{p}$. The initial term, $\ell_{pnp}$, is determined by measuring the difference between the predicted $\hat{p}$ and the PnP-estimated $\dot{p}$ 2D keypoint coordinates. Next, the PnP 2D keypoint coordinates are employed to generate a heatmap $\hat{h}$, which is then compared with the predicted heatmap $\hat{h}$ to establish the second term $\ell_{\dot{h}}$. The integration of these losses ensures the consistency of predicted heatmaps with the 3D keypoint locations of the spacecraft, even in the absence of ground-truth supervision.}
    \label{fig:heatmap-regularization-pipeline}
\end{figure}

Novel-view synthesis provides a strong learning objective to retrieve an accurate relative pose between two images, but does not impose any restriction on the heatmap response. This entails that if at test-time only the image reconstruction term $\ell_I$ is optimised, the heatmap responses would diverge to trivial solutions that are not consistent with the projection of the spacecraft keypoints $P_i$ over the image. To enforce solutions that are consistent with the structure of the 3D keypoints of the spacecraft, we introduce two loss terms (see Figure~\ref{fig:heatmap-regularization-pipeline}). The first term enforces that the maximum value of each heatmap channel is consistent with the set of possible projections by introducing pose information with a PnP-derived loss. Similarly as in~\cite{perez2023spacecraft}, we compare the estimated keypoint positions with those resulting of projecting the spacecraft keypoints $P_i$ onto the image, employing the PnP derived pose. First, the predicted image coordinates $\hat{p}_i$ are extracted from the predicted heatmap $\hat{h}$ with the expectation operation over the soft-max normalised heatmap, as performed in the integral pose regression framework~\cite{sun2018integral}. Then, the pose is retrieved with the backpropagatable PnP algorithm from~\cite{chen2020end}. This retrieved pose is employed over Equation~\ref{eq:perspective} to define the PnP estimated coordinates $\dot{p}_i$. Finally, the first term of the regularisation loss is derived by measuring the mean squared error between the expected image coordinates obtained by the heatmap $\hat{p}$ and the PnP derived coordinates $\dot{p}_i$:

\begin{equation}
    \ell_{pnp} =  \frac{1}{N} \sum_{i=1}^{N}  \left(\lVert \hat{p}_i-\dot{p}_i\rVert^2_2 +   \lVert \dot{p}_i - p_i\rVert^2_2 \right).\label{eq:loss_pnp}
\end{equation}

The second term of the Equation~\ref{eq:loss_pnp}, the one supervised by the ground-truth keypoint positions $p_i$ is added to regularise the output of the model and only optimised during the supervised stage. One drawback of $\ell_{pnp}$ is that only regulates where the expected maximum value of the heatmap is to be placed, however, the heatmaps should have only one response per channel and ideally have the same standard deviation as the one used for training. To enforce this behaviour we generate pseudo ground-truth heatmaps $\dot{h}$ with the estimated 2D keypoint positions obtained from PnP $\dot{p}$ in a differentiable manner employing the Kornia library~\cite{riba2020kornia}. Note that this loss term depends on the previous PnP term, and as such it cannot be evaluated independently.

\begin{equation}
    \ell_{\dot{h}} = \frac{1}{NWH} \Vert \hat{h} - \dot{h} \rVert_F^2. \label{eq:loss_heatmap_render}
\end{equation}

The proposed heatmap regularisation loss is then:

\begin{equation}
    \ell_{reg} =\ell_{pnp} + \ell_{\dot{h}}.
\end{equation}

\subsection{Overall solution} \label{sec:overall-solution}

\begin{figure*}
    \centering
    \includegraphics[width=0.9\textwidth]{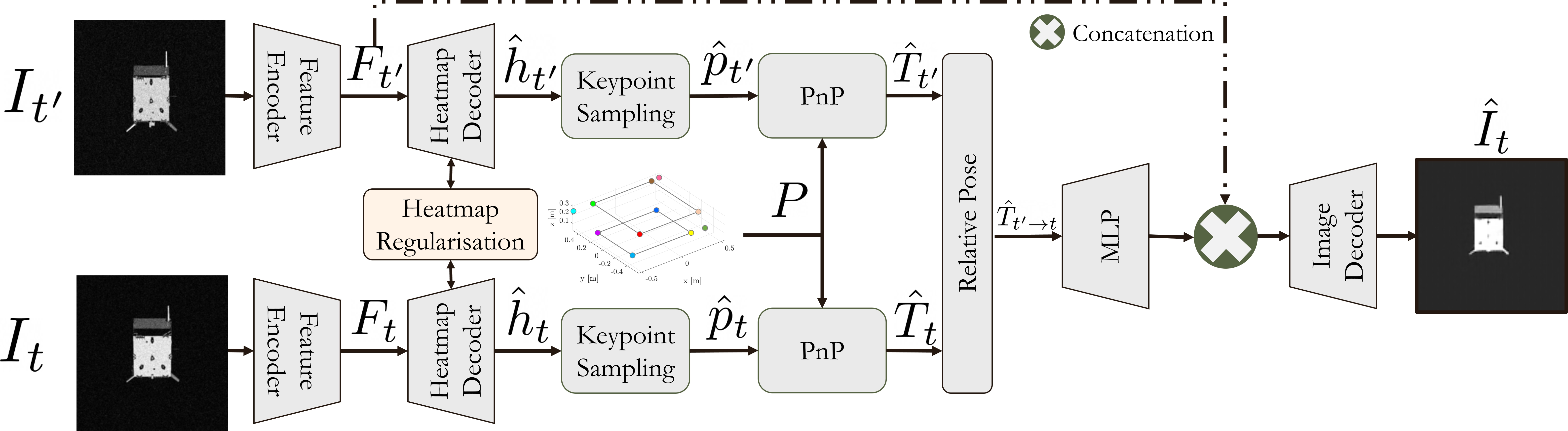}
    \caption{Final proposed architecture. The relative pose obtained from regressing the absolute poses of the spacecraft in two different time instants is used in conjunction with the image features of $I_{t'}$ to estimate the image $\hat{I}_t$. The difference between the estimated image $\hat{I}_t$ and the actual image $I_t$ are used as a self-supervised learning objective in conjunction with the heatmap regularisation loss.}
    \label{fig:fn_arch}
\end{figure*}

The complete solution scheme is depicted in Figure~\ref{fig:fn_arch}. Given two consecutive images $I_t$ and $I_{t'}$ of the target spacecraft, acquired at time instants $t$ and $t'$ and representing the spacecraft in the ground-truth poses $T_t$ and $T_{t'}$, each image is fed to a feature encoder to obtain a feature representation $F_t$ and $F_{t'}$, respectively; next, these feature representations are used by the pose estimation branch to obtain an estimate of the absolute pose of the spacecraft $\hat{T}_{t}$ and $\hat{T}_{t'}$ (see Section~\ref{seq:pose-estimation}). The spacecraft pose change between $t$ and $t'$, $T_{t\rightarrow t'}$, is used in combination with $F_t$ to obtain an estimate of the novel-view $\hat{I}_{t'}$ (see Section~\ref{seq:test-time adaptation-novel-view}). Inspired on self-supervised approaches that employ novel-view synthesis~\cite{godard2019digging}, we choose $t' = \{ t- \Delta, t+\Delta \}$ with $\Delta$ representing a scalar indicating the time difference between two temporally adjacent frames. Previous and next frames are included to encourage temporal consistency. The final training objective on the source domain is defined as:

\begin{equation}
    \ell_{train} = \ell^t_{h} + \ell^t_{reg}  + \sum_{t'} ( \ell^{t'}_{h} + \ell^{t'}_{reg} + \ell^{t'}_I).
\end{equation}

With the superindex indicating the time instant of the data employed to compute the loss. The heatmap loss $\ell_h$ and regularisation loss $\ell_{reg}$ are computed for all input images $t$ and $t'$, whereas the image reconsutrction loss $\ell_I$ is only computed for $t'$ as the image $I_t$ is used as the ground-truth view. During test-time adaptation on the target domain only the self-supervised losses are optimised:

\begin{equation}
    \ell_{tta} = \sum_{t'} ( \ell^{t'}_{reg} + \ell^{t'}_I).
\end{equation}

\section{Experiments}\label{seq:experiments}

In this section we present our experimental results. We assess the efficacy of our proposed approach by conducting experimental tests on datasets featuring both synthetically generated data and data captured under laboratory conditions (hardware-in-the-loop). This dual-environment testing enables us to train the models in a synthetic domain and evaluate them in a distinct, real-world domain. Such an approach allows us to quantify the performance decline caused by the domain gap and demonstrate how our proposed method mitigates this drop in performance. First we describe the experimental setup followed by the datasets and metrics employed. Next, we provide reference values of the standalone pose estimation methods and the improvements achieved with the proposed test-time adaptation. We continue by showing ablation studies on adaptation strategies and on the heatmap regularisation losses. We conclude the experiments by comparing our method against state-of-the-art methods and showing visual results.

\subsection{Experimental Setup}

We conduct our experiments using a ResNet-50 model from~\cite{xiao2018simple} for the feature encoder, the heatmap decoder and the image decoder. The image encoder consists of a two-layer fully connected network with 1024 neurons and softplus activations. Our framework is implemented in PyTorch~\cite{paszke2019pytorch} and Kornia~\cite{riba2020kornia}. We employ the Adam optimiser~\cite{kingma2014adam} with learning rate of 2.5e-4. We have decided to reduce the use of image augmentation to a minimum so that we can isolate its effects from the proposed test-time adaptation method. We just augment the training images by performing a Gaussian smoothing to mitigate the effects of noise and normalising them by the mean and standard deviation values of the training dataset.

\subsection{Datasets and Metrics}\label{sec:datasets-and-metrics}

\begin{figure}
    \centering
    \includegraphics[width=\columnwidth]{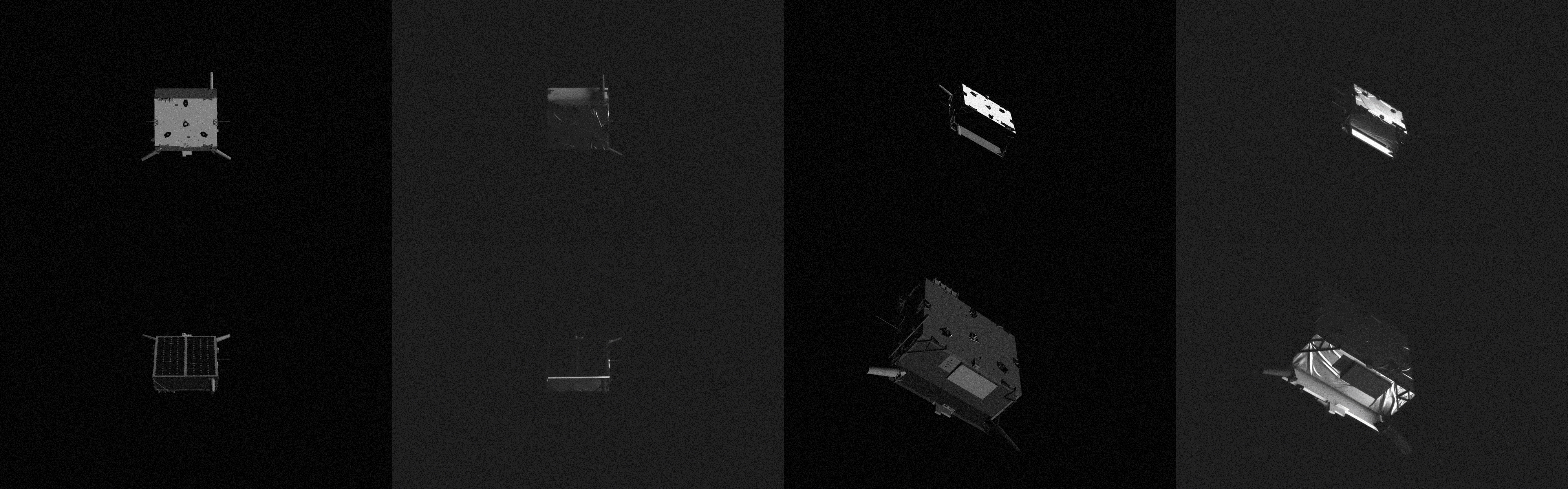}
    \caption{Example images for the SHIRT dataset. The two first columns represent the ROE1 dataset, the two last columns represent the ROE2 sequence. The first and third column are acquired in the synthetic domain, whereas the second and fourth are acquired in the Lightbox domain.}
    \label{fig:dataset-shirt}
\end{figure}

To train and evaluate our approach we employ the SPEED+ ~\cite{park2022speedplus} and the SHIRT~\cite{park2022adaptive} datasets. Both for SPEED+ and SHIRT we employ their standard training, validation, and test splits. Being the training and validation splits synthetically generated by a computer rendering software, and the test splits being captured in hardware-in-the-loop, laboratory conditions. The SPEED+ Dataset is comprised of 60,000 computer-generated training images of the Tango spacecraft with associated pose labels divided into 47,966 training images and 11,994 validation images). The test set is composed by 9,531 hardware-in-the-loop test images of a half-scale mock-up model acquired in two different illumination conditions: Lightbox and Sunlamp. With 6740 and 2791 images belonging to the Lightbox and Sunlamp domains respectively. The SHIRT Dataset extends this concept by generating sequential images, acquired every 5 seconds, of the target mock-up satellite in simulated rendezvous trajectories~\cite{park2022adaptive}. SHIRT provides two sequences: ROE1 and ROE2, represented in Figure~\ref{fig:dataset-shirt}. Each sequence contains 2,371 images and they are available in the Synthetic domain (used for training) and for the hardware-in-the-loop Lightbox domain (used for testing), with identical labels. In ROE1, the chaser spacecraft maintains a standard v-bar hold point with a fixed along-track separation, while the target rotates around one principal axis. On the other hand, in ROE2, the service gradually approaches the target, which is tumbling around two principal axes. For the SHIRT dataset we employ full set of synthetic images of ROE1 and ROE2 for training and the corresponding models, and the corresponding hardware-in-the-loop sets (Lightbox) for testing.

For evaluation, we employ the metrics from~\cite{park2023satellite}: the translation error $E_v$, defined as the Euclidean distance between a estimated translation vector $\hat{v}$ and the ground-truth translation vector $v$ : $E_v = \lVert \hat{v} - v \rVert_2$; and the orientation error $E_q$, computed as the rotation angle that aligns the estimated quaternion $\hat{q}$, and the corresponding ground-truth quaternion $q$: $E_q = 2 \cdot arccos(\left| \langle \hat{q},q\rangle\right|)$. Both errors are expressed in terms of scores: the translation score $S_v = E_v/\lVert v \rVert_2$, the orientation score $S_q = E_q$ and the total score $S= S_q + S_v$. Translation and orientation scores lower than the precision of the robotic arm employed to capture the hardware-in-the-loop domains ($2.173$x$10^{-3}$ and $0.169^\circ$) are set to zero{~\cite{park2023satellite}}.

\begin{figure}[h]
    \centering
    \includegraphics[width=\columnwidth]{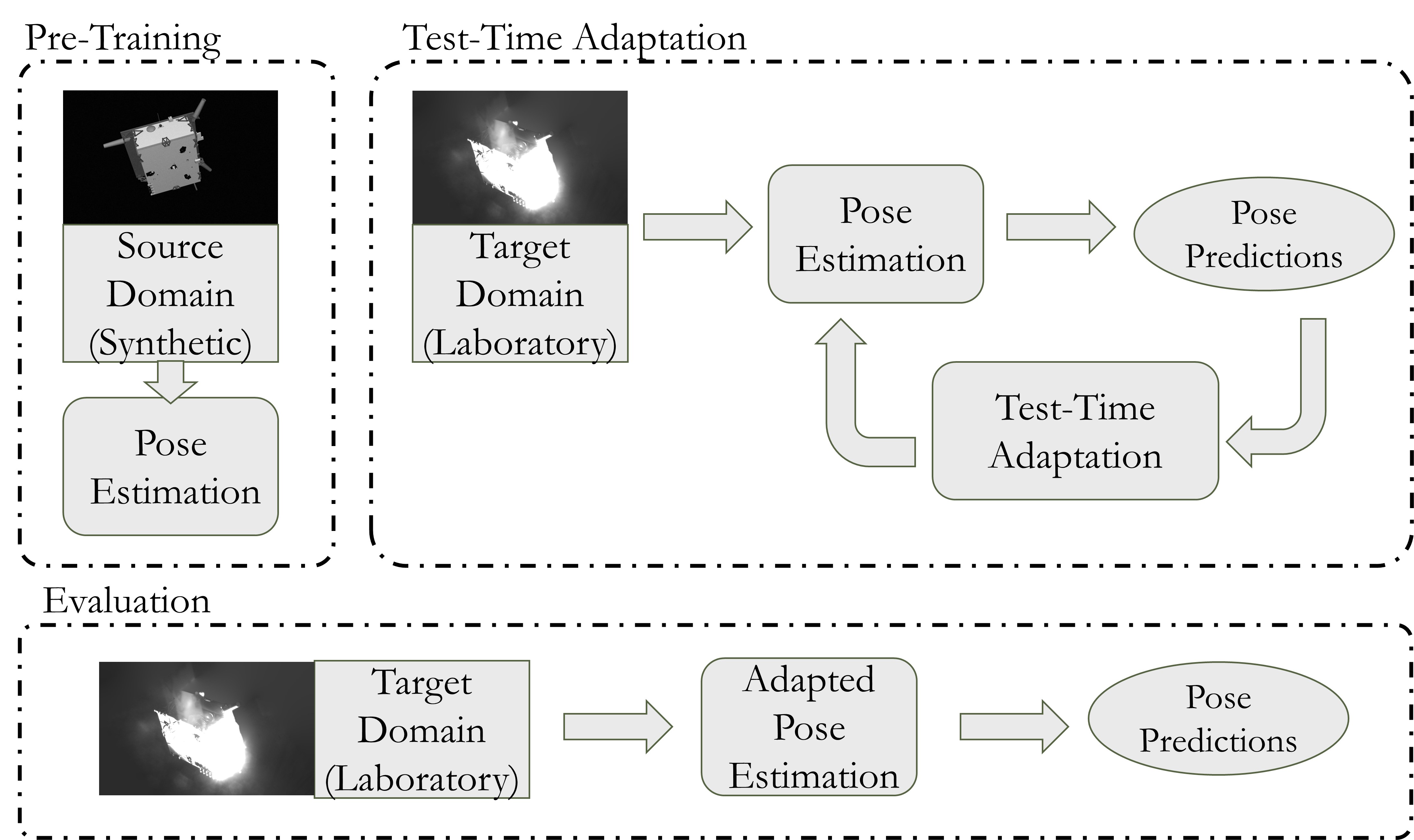}
    \caption{This flowchart outlines the experimental procedure, starting with pre-training the pose estimation model on a synthetic source domain, followed by test-time adaptation on a laboratory-captured target dataset, and concluding with the evaluation of the model's performance in the target domain.}
    \label{fig:experimental-flowchart}
\end{figure}

Figure~{\ref{fig:experimental-flowchart}} illustrates the experimental process we followed. Initially, we pre-train the pose estimation model using the synthetic source domain of the datasets. Subsequently, in order to enhance the pose estimation model, we apply test-time adaptation to the model on the target dataset, which consists of data captured under laboratory conditions. After adapting the model, we evaluate its final performance in the target domain.

\subsection{Baseline Performances}

As a reference, we provide in this section the performance values of the standalone pose estimation method described in Section~\ref{seq:pose-estimation}. We first report the results of training the method on either SHIRT or SPEED+ and then evaluating it on the Synthetic (S) and Lightbox (L) domains of SHIRT. We train SHIRT during 100 epochs and SPEED+ during 60 epochs, until their validation scores converge. For the evaluation, all layers of the model are frozen. The rationale behind these experiments is twofold: a) show the effect of the domain gap for the baseline method when trained on a large dataset (SPEED+) and when trained over the same sequence (SHIRT ROE1 or ROE2, Synthetic or Lightbox) but evaluated either on the same or on a different domain (Synthetic or Lightbox); b) provide the reference error bound by training and testing the proposed method on the same domain. 

The results of these experiments, evaluated in terms of the metrics defined in Section~\ref{seq:experiments}.~\ref{sec:datasets-and-metrics}, are reported in Table~\ref{tab:inter-domain-results}. For the ROE1 and ROE2 training experiments we can observe the effect of the domain gap for models trained and evaluated on different domains, with a higher error on the ROE2 trajectory likely caused by its higher complexity. This holds not only for models trained on the Synthetic domain and evaluated on the more complex domain Lightbox, but also we observe this behaviour for models trained on the Lightbox domain and evaluated on the Synthetic domain. The achieved scores are far from the reference error bound achieved by training and testing over the same dataset. For the SPEED+ dataset we can observe a similar behaviour, where the inter-domain error (trained on Synthetic and evaluated on Lightbox) is higher than the intra-domain error. Reference error bounds in the absence of domain gap are remarked in bold.

\begin{table}[!h]
\centering
\resizebox{1\columnwidth}{!}{%
\begin{tabular}{cc|cc|ccc|cc}
\multicolumn{2}{c|}{Train}                                        & \multicolumn{2}{c|}{Test}                      & $S_{v}$ & $S_{r}$ & $S$                                                        &  $E_v [m]$                & $E_r[\circ]$             \\ \hline
\multicolumn{1}{c|}{\multirow{4}{*}{ROE1}}   & \multirow{2}{*}{S} & \multicolumn{1}{c|}{\multirow{4}{*}{ROE1}} & S & \textbf{0.011}   & \textbf{0.020}   & \textbf{0.031}                           &  \textbf{0.083}           &  \textbf{1.173}           \\
\multicolumn{1}{c|}{}                        &                    & \multicolumn{1}{c|}{}                      & L & 0.072   & 0.542   & 0.614                                                      &  0.548                    &  31.050                   \\ \cline{2-2} \cline{4-9}  
\multicolumn{1}{c|}{}                        & \multirow{2}{*}{L} & \multicolumn{1}{c|}{}                      & S & 0.396   & 1.205   & 1.601                                                      &  3.034                    &  69.069                   \\
\multicolumn{1}{c|}{}                        &                    & \multicolumn{1}{c|}{}                      & L & \textbf{0.011}     & \textbf{0.021}   & \textbf{0.032}                            & \textbf{0.083}            &  \textbf{1.219}           \\ \hline  
\multicolumn{1}{c|}{\multirow{4}{*}{ROE2}}   & \multirow{2}{*}{S} & \multicolumn{1}{c|}{\multirow{4}{*}{ROE2}} & S & \textbf{0.010}   & \textbf{0.014}   & \textbf{0.024}                           & \textbf{0.057}            &  \textbf{0.776}           \\
\multicolumn{1}{c|}{}                        &                    & \multicolumn{1}{c|}{}                      & L & 0.526   & 2.131   & 2.660                                                      &  2.896                    &  122.280                  \\ \cline{2-2} \cline{4-9} 
\multicolumn{1}{c|}{}                        & \multirow{2}{*}{L} & \multicolumn{1}{c|}{}                      & S & 0.483   & 1.810    & 2.293                                                     &  2.581                    &  103.684                  \\
\multicolumn{1}{c|}{}                        &                    & \multicolumn{1}{c|}{}                      & L & \textbf{0.010}   & \textbf{0.012}   & \textbf{0.022}                           & \textbf{0.053}            &  \textbf{0.674}           \\ \hline
\multicolumn{1}{c|}{\multirow{4}{*}{SPEED+}} & \multirow{4}{*}{S} & \multicolumn{1}{c|}{\multirow{2}{*}{ROE1}} & S & \textbf{0.302}  & \textbf{1.008}   &\textbf{1.310}                             & \textbf{2.316}            & \textbf{57.766}           \\
\multicolumn{1}{c|}{}                        &                    & \multicolumn{1}{c|}{}                      & L & 0.676   & 2.100   & 2.777                                                      & 5.157                     & 120.343                   \\ \cline{3-9} 
\multicolumn{1}{c|}{}                        &                    & \multicolumn{1}{c|}{\multirow{2}{*}{ROE2}} & S & \textbf{0.151}   & \textbf{0.631}   & \textbf{0.782}                           & \textbf{0.821}            & \textbf{36.152}           \\
\multicolumn{1}{c|}{}                        &                    & \multicolumn{1}{c|}{}                      & L & 0.506   & 1.936   & 2.442                                                      & 2.627                     & 110.946                    
\end{tabular}%
}
\caption{Baseline pose estimation results. Scores achieved by training the standalone pose estimation method on the train column and evaluating on the test column. S indicates Synthetic domain and L indicates Lightbox domain. Reference error bounds for the same sequence in the absence of domain gap are remarked in bold.}
\label{tab:inter-domain-results}
\end{table}

\subsection{Test-Time Adaptation Performance} \label{seq:experiments-test-time adaptation}

Our proposed test-time adaptation method requires a sequence of images to predict the appearance of an image based on two pose estimates and on a time-adjacent image. Hence, we train and evaluate our method on the SHIRT, as it provides rendezvous trajectories in two domains: the Synthetic one and the Lightbox one. We first train the overall method described in Section~\ref{sec:overall-solution} for 100 epochs over the SHIRT Synthetic domain, optimising both supervised and self-supervised losses. Then, we perform the test-time adaptation during 20 epochs by optimising only the self-supervised objectives on the Lightbox domain. We run the test-time adaptation experiments five separate times and report the mean and standard deviation of the metrics defined in Section~\ref{seq:experiments}.~\ref{sec:datasets-and-metrics}.

\begin{table}[!h]
\centering
\resizebox{1\columnwidth}{!}{%
\begin{tabular}{c|c|ccc|cc}
Adaptation                  & Test & $S_{v}$             & $S_{r}$           & $S$               &  $E_v [m]$       & $E_r[\circ]$  \\ \hline
\multirow{2}{*}{No}  & ROE1 & 0.072               & 0.542             & 0.614                    &  0.548           & 31.050        \\
                     & ROE2 & 0.526               & 2.131             & 2.660                    &  2.896           & 122.280       \\ \hline
\multirow{2}{*}{Yes} & ROE1 & 0.019 $\pm$ 0.001 & 0.063 $\pm$ 0.002 & 0.083 $\pm$ 0.002          &  0.21 $\pm$  0.12    &  4 $\pm$  5         \\
                     & ROE2 & 0.041 $\pm$ 0.002 & 0.159 $\pm$ 0.012 & 0.200 $\pm$ 0.014          &  0.22 $\pm$  0.42    & 8 $\pm$ 23

\end{tabular}%
}
\caption{Results of training on the Synthetic domain of the SHIRT dataset and testing on the Lightbox domain. The results are shown for the standalone method (no adaptation) and for the proposed test-time adaptation method. }
\label{tab:inter-domain-results-w-test-time adaptation}
\end{table}

It can be observed that the proposed test-time adaptation approach reduces the overall error by a large margin just requiring the input images of the test-sequence: 0.083 vs 0.614 for the case of ROE1 and 0.2 vs 2.66 for the case of ROE2. During test-time adaptation we optimise all the networks at a reduced learning rate to increase stability during training and we sample one of each 5 images. An extended analysis on these choices is provided in Section~\ref{seq:freezing-and-deltas}.

\subsection{Ablation Studies}
This section presents a study on the best configuration for the proposed test-time adaptation stage, along with an incremental analysis on the effect of the heatmap regularisation losses on the pose estimation performance. To present our findings, we have chosen to use pose scores while omitting the $E_v$ and $E_r$ metrics. This decision is made to enhance clarity due to the extensive nature of the results.

\subsubsection{Freezing and Deltas}\label{seq:freezing-and-deltas}

The proposed test-time adaptation approach includes two design parameters: 1) the choice of freezing or enabling updates of the novel-view synthesis stage; and 2) the learning rate to employ during adaptation. The novel-view estimation pipeline acts as a referee during the test-time adaptation, providing an indicator of the quality of the pose estimation. Ideally, this pipeline should remain fixed to avoid changes in the encoder that only favour the novel-view estimation. However, as the adaptation progresses, the feature vector that encodes the heatmap information that is used for such estimation will change, incurring into a situation where displacements of the feature vector that improve the heatmap estimation may reduce the view estimation performance. We study the effect of fixing and updating the novel-view estimation pipeline for different configurations of learning rates. Additionally, the study analyses the sampling effect, i.e. that of increasing the time distance between two target and source images. This sampling value is represented by~$\Delta$.

\begin{figure*}[!h]
    \centering
    \includegraphics[width=0.8\textwidth]{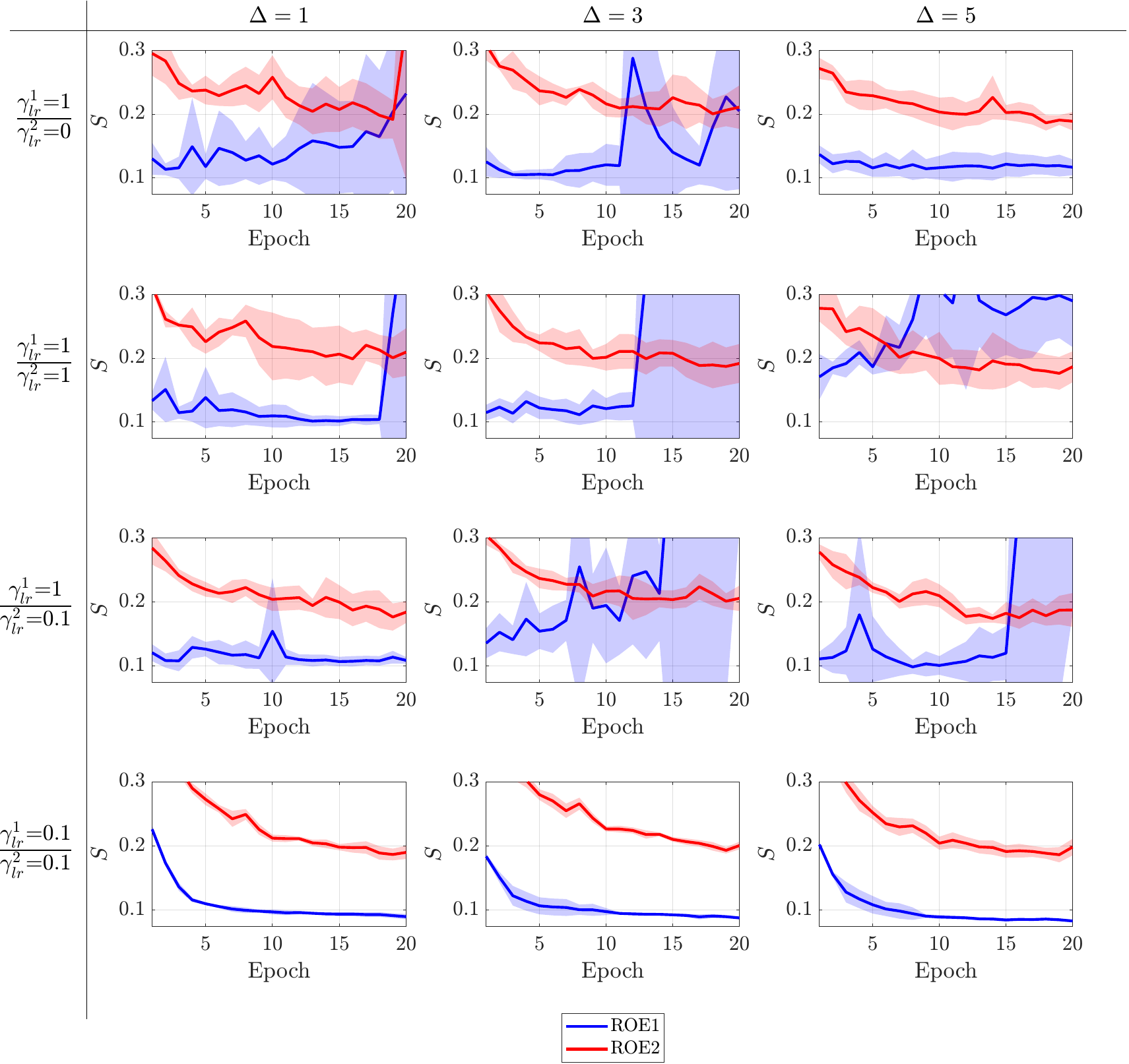}
    \caption{Study on the effect of sampling ($\Delta$) and of the update strategy for the novel-view-synthesis stage on the test-time adaptation performance expressed in terms of $S$. Each model has been pretrained on the corresponding $\Delta$ value prior to adaptation.}
    \label{fig:freezing-and-delta}
\end{figure*}

The results of these experiments are represented in Figure~\ref{fig:freezing-and-delta} in terms of the overall pose score $S$. Each column corresponds to a different $\Delta = 1, 3, 5$ frames that is equivalent to 5, 15, and 25 seconds of time difference between images, whereas each row represents a different updating strategy: $\gamma^1_{lr}$  represents the weight applied to the original learning rate of the feature encoder and heatmap decoder; and $\gamma^2_{lr}$ represents the weight applied to the original learning rate of the novel-view synthesis stage. According to their values: 1) $\gamma^1_{lr}=1$ and $\gamma^2_{lr}=0$ indicates that the novel-view synthesis module remains fixed and the pose estimation pipeline uses the same learning rate for adaptation as the one used during training; 2)  $\gamma^1_{lr}=1$ and $\gamma^2_{lr}=1$  indicates that the test-time adaptation stages are updated with the same learning rate as used during training; 3)  $\gamma^1_{lr}=1$ and $\gamma^2_{lr}=0.1$ are set with the rationale of reducing the learning rate of the image predictor only to let the novel-view synthesis model to adapt to changes in feature distribution but not to have a learning power large enough to optimise only the view estimation loss; and 4) $\gamma^1_{lr}=0.1$ and $\gamma^2_{lr}=0.1$ are set to train all modules at a smaller learning rate. The results are obtained by performing each test-time adaptation configuration during 20 epochs in 5 different runs to obtain the mean error represented with the continuous line, and the standard deviation metrics represented with shaded areas.

We can observe that better performances are achieved for the ROE1 dataset across all experiments, likely influenced by the lower complexity of the trajectory compared to ROE2. Setting $\gamma^1_{lr}=\gamma^2_{lr}=0.1$ yields the lowest deviation for all deltas and training epochs, at the cost of requiring more iterations to achieve a lower error. For $\gamma^2_{lr} = 0$ and $\gamma^2_{lr}=0.1$ similar results are obtained: the error after optimising the test-time adaptation is similar across deltas but tends to diverge, likely because the novel-view synthesis cannot adapt to the changes in feature distribution. Optimising with $\gamma^1_{lr} = \gamma^2_{lr} = 1$ does not provide consistent results, tending to diverge in the ROE1 sequence or to generate large deviation errors for ROE2. Regarding the $\Delta$ value we cannot observe a consistent effect across experiments except for the $\gamma^1_{lr} = \gamma^2_{lr} = 0.1$ case, where it shows a trade-off between the mean final error and the stability during training. The mean final error slightly decreases for larger values of $\Delta$ but decreases the stability of the method. 
These experiments show that a trade-off between stability and processing time can be achieved: allowing for slower convergence provides more stable results but requires more computational resources, whereas quicker optimisation methods can be employed at the cost of larger uncertainty on the final performance. Table~\ref{tab:1-epoch-test-time adaptation} provides a comparison on the results of running the proposed test-time adaptation method during one epoch and the final results achieved by optimising with $\gamma^1_{lr}=\gamma^2_{lr}=0.1$ during 20 epochs, showcasing the observed trade-off between training time and performance.
\begin{table*}[!h]
\centering
\resizebox{1\textwidth}{!}{%
\begin{tabular}{ccccccccc}
\multicolumn{1}{l}{}                      & \multicolumn{1}{l}{}                      & \multicolumn{1}{l|}{}         & \multicolumn{3}{c|}{ROE 1}                                                     & \multicolumn{3}{c}{ROE 2}                                                     \\ \hline
$\gamma^1_{lr}$                           & \multicolumn{1}{c|}{$\gamma^2_{lr}$}      & \multicolumn{1}{c|}{$\Delta$} & $S_{v}$           & $S_{r}$           & \multicolumn{1}{c|}{$S$}               & $S_{v}$           & $S_{r}$           & $S$                                   \\ \hline
\multicolumn{9}{c}{\textbf{1 Epoch}}                                                                                                                                                                                                                                                   \\ \hline
\multicolumn{1}{c|}{\multirow{3}{*}{1}}   & \multicolumn{1}{c|}{\multirow{3}{*}{0}}   & \multicolumn{1}{c|}{1}        & 0.026 $\pm$ 0.002 & 0.088 $\pm$ 0.008 & \multicolumn{1}{c|}{0.114 $\pm$ 0.009} & 0.056 $\pm$ 0.007 & 0.228 $\pm$ 0.03  & 0.284 $\pm$ 0.036                     \\
\multicolumn{1}{c|}{}                     & \multicolumn{1}{c|}{}                     & \multicolumn{1}{c|}{3}        & 0.023 $\pm$ 0.002 & 0.09 $\pm$ 0.011  & \multicolumn{1}{c|}{0.113 $\pm$ 0.013} & 0.056 $\pm$ 0.003 & 0.220 $\pm$ 0.003 & 0.275 $\pm$ 0.006                     \\
\multicolumn{1}{c|}{}                     & \multicolumn{1}{c|}{}                     & \multicolumn{1}{c|}{5}        & 0.027 $\pm$ 0.001 & 0.095 $\pm$ 0.014 & \multicolumn{1}{c|}{0.122 $\pm$ 0.015} & 0.054 $\pm$ 0.002 & 0.210 $\pm$ 0.012 & 0.264 $\pm$ 0.012                     \\ \hline
\multicolumn{1}{c|}{\multirow{3}{*}{1}}   & \multicolumn{1}{c|}{\multirow{3}{*}{1}}   & \multicolumn{1}{c|}{1}        & 0.028 $\pm$ 0.002 & 0.123 $\pm$ 0.05  & \multicolumn{1}{c|}{0.151 $\pm$ 0.051} & 0.053 $\pm$ 0.004 & 0.215 $\pm$ 0.007 & 0.268 $\pm$ 0.010                     \\
\multicolumn{1}{c|}{}                     & \multicolumn{1}{c|}{}                     & \multicolumn{1}{c|}{3}        & 0.025 $\pm$ 0.003 & 0.095 $\pm$ 0.012 & \multicolumn{1}{c|}{0.12 $\pm$ 0.015}  & 0.055 $\pm$ 0.005 & 0.22 $\pm$ 0.030   & \multicolumn{1}{l}{0.275 $\pm$ 0.034} \\
\multicolumn{1}{c|}{}                     & \multicolumn{1}{c|}{}                     & \multicolumn{1}{c|}{5}        & 0.032 $\pm$ 0.004 & 0.153 $\pm$ 0.013 & \multicolumn{1}{c|}{0.185 $\pm$ 0.010}  & 0.058 $\pm$ 0.007 & 0.22 $\pm$ 0.031  & 0.278 $\pm$ 0.037                     \\ \hline
\multicolumn{1}{c|}{\multirow{3}{*}{1}}   & \multicolumn{1}{c|}{\multirow{3}{*}{0.1}} & \multicolumn{1}{c|}{1}        & 0.026 $\pm$ 0.002 & 0.083 $\pm$ 0.009 & \multicolumn{1}{c|}{0.109 $\pm$ 0.011} & 0.051 $\pm$ 0.004 & 0.213 $\pm$ 0.010  & 0.264 $\pm$ 0.017                     \\
\multicolumn{1}{c|}{}                     & \multicolumn{1}{c|}{}                     & \multicolumn{1}{c|}{3}        & 0.026 $\pm$ 0.002 & 0.127 $\pm$ 0.028 & \multicolumn{1}{c|}{0.153 $\pm$ 0.03}  & 0.056 $\pm$ 0.003 & 0.228 $\pm$ 0.007 & 0.284 $\pm$ 0.007                     \\
\multicolumn{1}{c|}{}                     & \multicolumn{1}{c|}{}                     & \multicolumn{1}{c|}{5}        & 0.026 $\pm$ 0.003 & 0.087 $\pm$ 0.022 & \multicolumn{1}{c|}{0.113 $\pm$ 0.024} & 0.053 $\pm$ 0.003 & 0.205 $\pm$ 0.021 & 0.258 $\pm$ 0.020                      \\ \hline
\multicolumn{1}{c|}{\multirow{3}{*}{0.1}} & \multicolumn{1}{c|}{\multirow{3}{*}{0.1}} & \multicolumn{1}{c|}{1}        & 0.026 $\pm$ 0.001 & 0.148 $\pm$ 0.005 & \multicolumn{1}{c|}{0.173 $\pm$ 0.005} & 0.074 $\pm$ 0.002 & 0.282 $\pm$ 0.005 & 0.356 $\pm$ 0.005                     \\
\multicolumn{1}{c|}{}                     & \multicolumn{1}{c|}{}                     & \multicolumn{1}{c|}{3}        & 0.021 $\pm$ 0.001 & 0.130 $\pm$ 0.008 & \multicolumn{1}{c|}{0.151 $\pm$ 0.008} & 0.074 $\pm$ 0.003 & 0.284 $\pm$ 0.01  & 0.358 $\pm$ 0.015                     \\
\multicolumn{1}{c|}{}                     & \multicolumn{1}{c|}{}                     & \multicolumn{1}{c|}{5}        & 0.023 $\pm$ 0.001 & 0.133 $\pm$ 0.004 & \multicolumn{1}{c|}{0.156 $\pm$ 0.004} & 0.068 $\pm$ 0.006 & 0.253 $\pm$ 0.02  & 0.321 $\pm$ 0.025                     \\ \hline
\multicolumn{9}{c}{\textbf{20 Epochs}}                                                                                                                                                                                                                                                 \\ \hline
\multicolumn{1}{c|}{\multirow{3}{*}{0.1}} & \multicolumn{1}{c|}{\multirow{3}{*}{0.1}} & \multicolumn{1}{c|}{1}        & 0.020 $\pm$ 0.001  & 0.070 $\pm$ 0.004  & \multicolumn{1}{c|}{0.090 $\pm$ 0.004}  & 0.039 $\pm$ 0.002 & 0.151 $\pm$ 0.009 & 0.190 $\pm$ 0.010                       \\
\multicolumn{1}{c|}{}                     & \multicolumn{1}{c|}{}                     & \multicolumn{1}{c|}{3}        & 0.018 $\pm$ 0.001  & 0.069 $\pm$ 0.002 & \multicolumn{1}{c|}{0.088 $\pm$ 0.002} & 0.041 $\pm$ 0.002 & 0.160 $\pm$ 0.005  & 0.201 $\pm$ 0.006                     \\
\multicolumn{1}{c|}{}                     & \multicolumn{1}{c|}{}                     & \multicolumn{1}{c|}{5}        & 0.019 $\pm$ 0.001 & 0.063 $\pm$ 0.002 & \multicolumn{1}{c|}{0.083 $\pm$ 0.002} & 0.041 $\pm$ 0.002 & 0.159 $\pm$ 0.012 & 0.200 $\pm$ 0.014                      
\end{tabular}%
}
\caption{Mean results achieved after running the proposed test-time adaptation method in all configurations for 1 epoch, and mean results obtained after running the $\gamma^2_{lr} = 0$ and $\gamma^2_{lr}=0.1$ method for 20 epochs.}
\label{tab:1-epoch-test-time adaptation}
\end{table*}

\subsubsection{Regularisation Losses}

\begin{figure}
    \centering
    \includegraphics[width=\columnwidth]{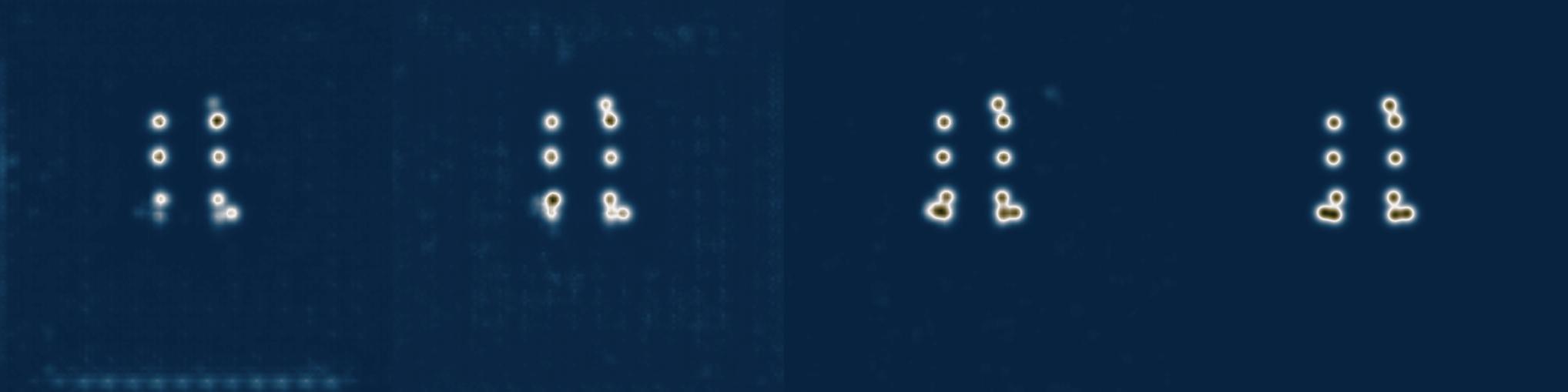}
    \caption{Effect of heatmap regularisation losses. From left to right the heatmap results of using $\ell_I$ for adaptation, and increasingly adding $\ell_{pnp}$ and $ \ell_{\dot{h}}$. Rightmost represents the ground-truth heatmap. We can observe artefacts in the lower part of the heatmap corresponding to $\ell_I$ and that the heatmap quality increases with the inclusion of the proposed regularisation losses.}
    \label{fig:heatmap-regularization}
\end{figure}

Table~\ref{tab:ablation-losses} compiles for ROE1 the effect of the regularization losses on the scores and Figure~\ref{fig:heatmap-regularization} its effect on the heatmap response. In the absence of regularisation losses, the keypoint heatmaps can diverge to the ones that explain the best relative pose between images but are not consistent with the inherent spacecraft keypoint structure, especially for higher learning rates. Integrating the $\ell_{pnp}$ loss greatly improves the results; however, it only imposes restrictions on the location of the maximum value of the heatmap, not affecting the modality of the heatmap. The best results are achieved when the pseudo ground-truth heatmap $\ell_{\dot{h}}$ is added, obtaining a pose score of 0.083 vs 0.121.  Note that the effect of $\ell_{\dot{h}}$ cannot be measured independently from $\ell_{pnp}$ as a pose estimate from PnP is required to generate the rendered heatmap.

\begin{table}[!h]
\centering
\resizebox{0.8\columnwidth}{!}{%
\begin{tabular}{l|lll}
Losses                          & \multicolumn{1}{c}{$S_{v}$} & \multicolumn{1}{c}{$S_{r}$} & \multicolumn{1}{c}{$S$} \\ \hline
$\ell_I$                                 & 0.035 $\pm$ 0.011            & 0.176 $\pm$ 0.037            & 0.211 $\pm$ 0.011        \\ \hline
$\ell_I + \ell_{pnp}$                    & 0.034 $\pm$ 0.008            & 0.086 $\pm$ 0.024            & 0.121 $\pm$ 0.029       \\ \hline
$\ell_I + \ell_{pnp} +   \ell_{\dot{h}}$ & 0.019 $\pm$ 0.001            & 0.063 $\pm$ 0.002            & 0.083 $\pm$ 0.002     
\end{tabular}%
}
\caption{Ablation study on the heatmap regularisation losses for the ROE1 sequence. Adding a guiding loss that encapsulates the expected heatmap response $\ell_{\dot{h}}$ improves the results over just considering the keypoint coordinates $\ell_{pnp}$.}
\label{tab:ablation-losses}
\end{table}

\subsection{Comparison With The State Of The Art}

In Table~\ref{tab:sota-comparison} we provide numerical comparison with state-of-the-art methods evaluated over SHIRT. We follow previous literature and report the final metrics in terms of $E_v$ and $E_r$. Additionally we compare the proposed method with a simple test-time adaptation method based on adapting the statistics of the Batch Normalization layers~\cite{li2018adaptive}. This experiment is set with the rationale that the object and its trajectory do not change between domains, and hence the features learnt in the Synthetic domain, while useful, might be displaced due to the different intensity range of the images.

Performing adaptation of the Batch Normalization layers as proposed by~\cite{li2018adaptive} helps reduce the overall error compared to just evaluating the model; however, the features learnt from the Synthetic domain are not fully transferable to those in the Lightbox domain. Regarding SPNv2~\cite{park2022robust,park2022adaptive} we report the results of the $E$ head for the translation error and those of the $H$ head for the orientation error, as these are the best corresponding values for ROE1 and ROE2. As SPNv2 is trained over the SPEED+ dataset (40k images) and our method over a much smaller one, SHIRT (2k images), but with the same observed poses, a direct comparison is not applicable. However the comparison shows the potential of our method which, without the use of any data and style augmentations, improves the overall pose scores of SPNv2 by a factor of 3~\cite{park2022robust}, enhancing the orientation estimation over ROE1 and achieving similar performance over ROE2. Our results further improve by introducing just the data augmentation pipeline of SPNv2, without including the style transfer augmentation. Although this is beyond the scope of this work, we assume that combining our method with SPNv2 or testing other architectures trained on more extensive and diverse sequences could lead to improved outcomes on the SHIRT dataset and on other sources of sequential data.

\begin{table}[]
\resizebox{\columnwidth}{!}{%
\begin{tabular}{c|cc|cc}
                     & \multicolumn{2}{c|}{ROE 1}     & \multicolumn{2}{c}{ROE 2}       \\ \cline{2-5} 
Method               & $E_v [m]$       & $E_r[\circ]$ & $E_v [m]$        & $E_r[\circ]$ \\ \hline
BN Adaptation~\cite{li2018adaptive}                         & $0.31 \pm 0.40$ & $16 \pm 31$  & $0.83 \pm 0.95$  & $51 \pm  52$ \\ \hline
SPNv2~\cite{park2022robust,park2022adaptive}                & $0.18 \pm 0.15$ & $17 \pm 42$  & $0.10 \pm 0.11$  & $5 \pm 14$   \\ \hline
Ours                                                        & $0.21 \pm 0.12$ & $4 \pm 5$    & $0.22 \pm 0.42$  & $8 \pm 23$  \\ 
Ours + Augmentation                                         & $0.11 \pm 0.08$ & $3 \pm 3$    & $0.09 \pm 0.23$  & $5 \pm 20$  
\end{tabular}%
}\caption{Comparison with other image-based state-of-the-art methods. We report our method as baseline and including data augmentation.}
\label{tab:sota-comparison}
\end{table}

\subsection{Visual Results}

\begin{figure}[ht]
    \centering
    \includegraphics[width=1\columnwidth]{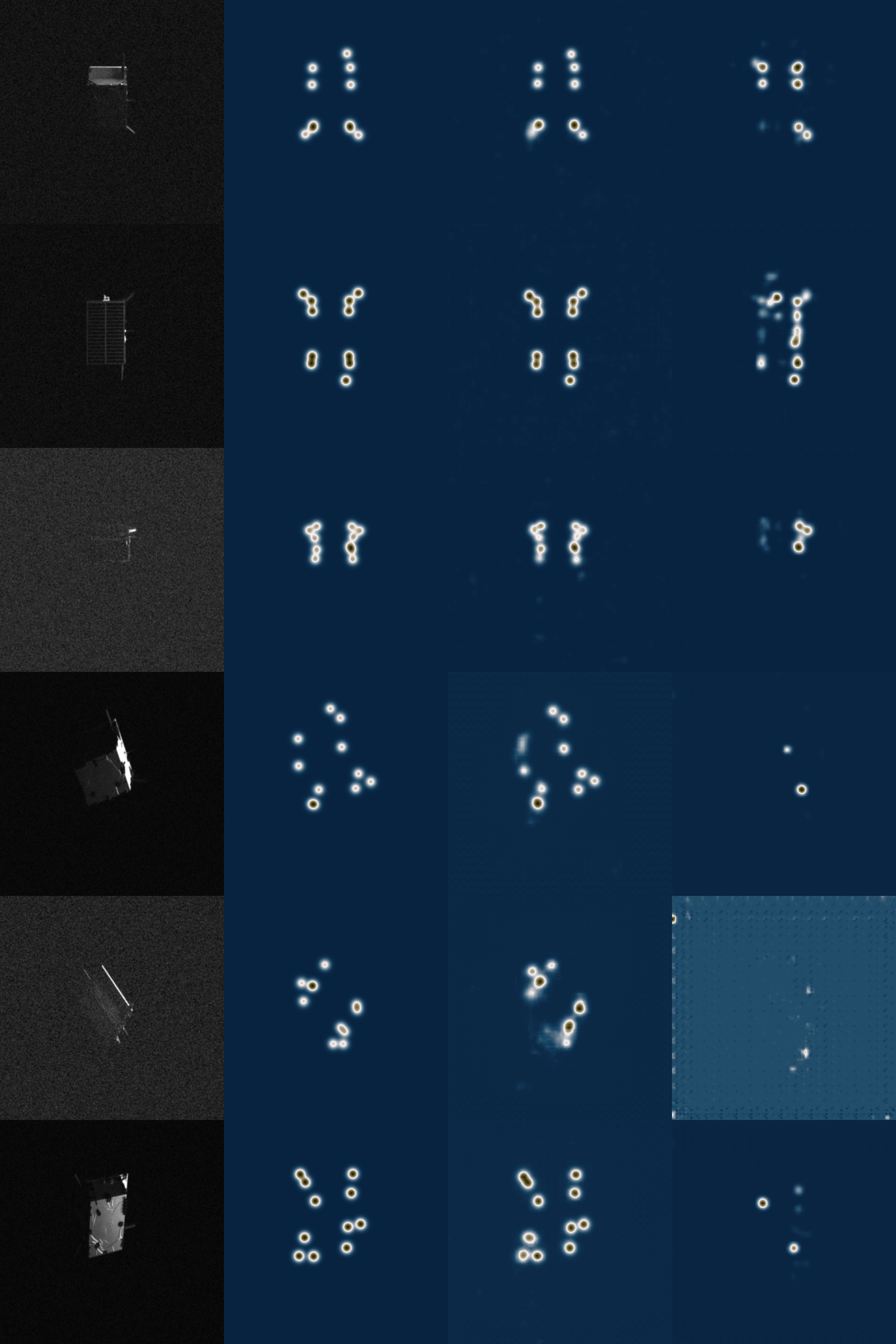}
    \caption{The columns, from left to right, represent: the input image of the target spacecraft, the ground-truth heatmap, the estimated heatmap obtained using the proposed test-time adaptation method, and the estimated heatmap before applying the adaptation. The first three rows showcase image examples from the ROE1 sequence, while the last three rows showcase the ROE2 sequence.}
    \label{fig:heatmap-examples}
\end{figure}

 Figure~\ref{fig:heatmap-examples} shows example images of the proposed test-time adaptation method. Similar to the numerical results, these examples illustrate how our proposal helps improve the heatmap estimation, which translates in a better pose estimation. Adaptation performs better over the ROE1 sequences than over the ROE2 ones, likely due to the higher complexity of the latter.

\begin{figure}[ht]
    \centering
    \includegraphics[width=1\columnwidth]{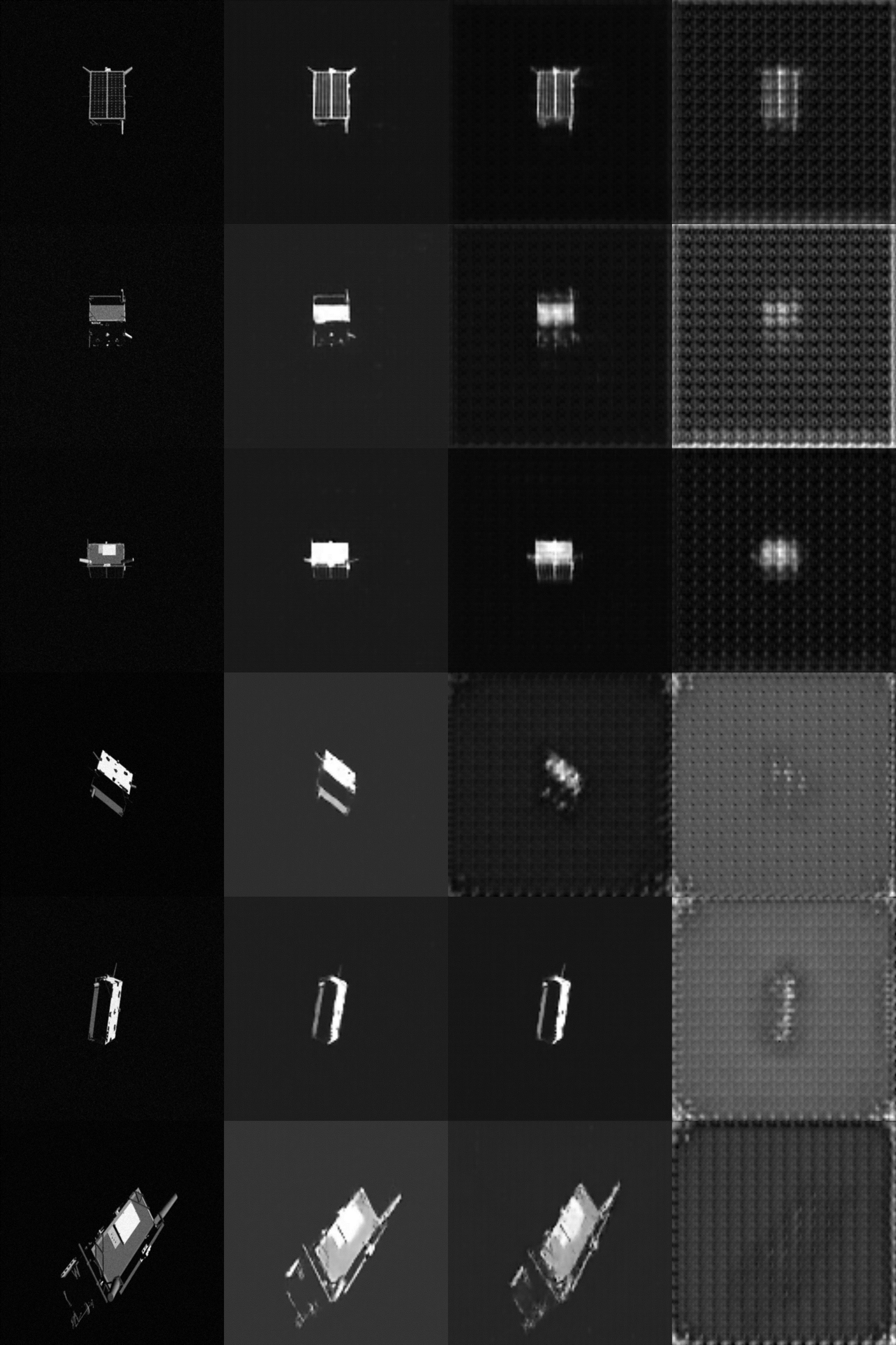}
    \caption{The columns, from left to right, represent: the input image of the target spacecraft, the reconstructed image without pose error ($\beta = 0$), the reconstructed image with a pose error of $\beta = 10$, and the reconstructed image with a pose error of $\beta = 100$. The first three rows showcase image examples from the ROE1 sequence, while the last three rows showcase the ROE2 sequence.}
    \label{fig:reconstruction-example}
\end{figure}

Finally, in Figure~\ref{fig:reconstruction-example} we represent example results of the novel-view synthesis module with increasingly higher pose errors in the input. The pose errors are introduced by adding uniform noise drawn from $[0,1]$ weighted by $\beta = 0, 10, 100$. We can observe how the quality of the estimated view is reduced as the error increases, showing artifacts for  $\beta = 10$ and noise for $\beta = 100$. The novel-view synthesis stage is designed to quantify the goodness of the pose estimate, not to render an image of the spacecraft in any input pose; this explains the presence of noise for pose transitions that are not observed during training. Also, we can observe that some examples are more affected by noise than others, which suggests that some mechanism to ensure consistent degradation of the image with increasingly pose errors would be beneficial for the method.

\section{Conclusions and Future Work}\label{seq:conclusions}

In this work we have presented a novel method to enhance keypoint-based spacecraft pose estimation on new domains. In particular, our presented contributions are:
\begin{itemize}
    \item A novel-view synthesis-based test-time adaptation method to enhance keypoint- based spacecraft pose estimation, that employs image sequence redundancy commonly available in this domain. We have shown through experiments and results that our approach improves the baseline pose estimation method and achieves performance that pair with state-of-the-art techniques that evaluate on the same datasets.
    \item A new set of heatmap regularisation losses, designed to embed the structure of spacecraft keypoints into self-supervised losses for keypoint learning.
\end{itemize}

As for future improvements, we suggest exploring methods to scale the ground-truth heatmaps based on the distance to the spacecraft, to ensure the same area is covered independently of the pose. Additionally, updating our novel-view synthesis pipeline with advanced elements like differentiable renderers or style transfer techniques could be beneficial. Another potential area to investigate is the performance of our proposed method when applied to different CNN or Transformer architectures~{\cite{wang2022revisiting}}. Another potential research direction is to enable the application in real operational scenarios, including adapting the model on few test data and measuring the performance on large test sequences, and methods to efficiently adapt the models onboard.

\textbf{Acknowledgements.}  This work is supported by Comunidad Aut\'onoma de Madrid (Spain) under the Grant IND2020/TIC-17515 and the HVD (PID2021-125051OB-I00) project funded by the Ministerio de Ciencia e Innovación of the Spanish Government.

\bibliographystyle{ieeetr}
\bibliography{references.bib}

\begin{IEEEbiography}[{\includegraphics[width=1in,height=1.25in,clip,keepaspectratio]{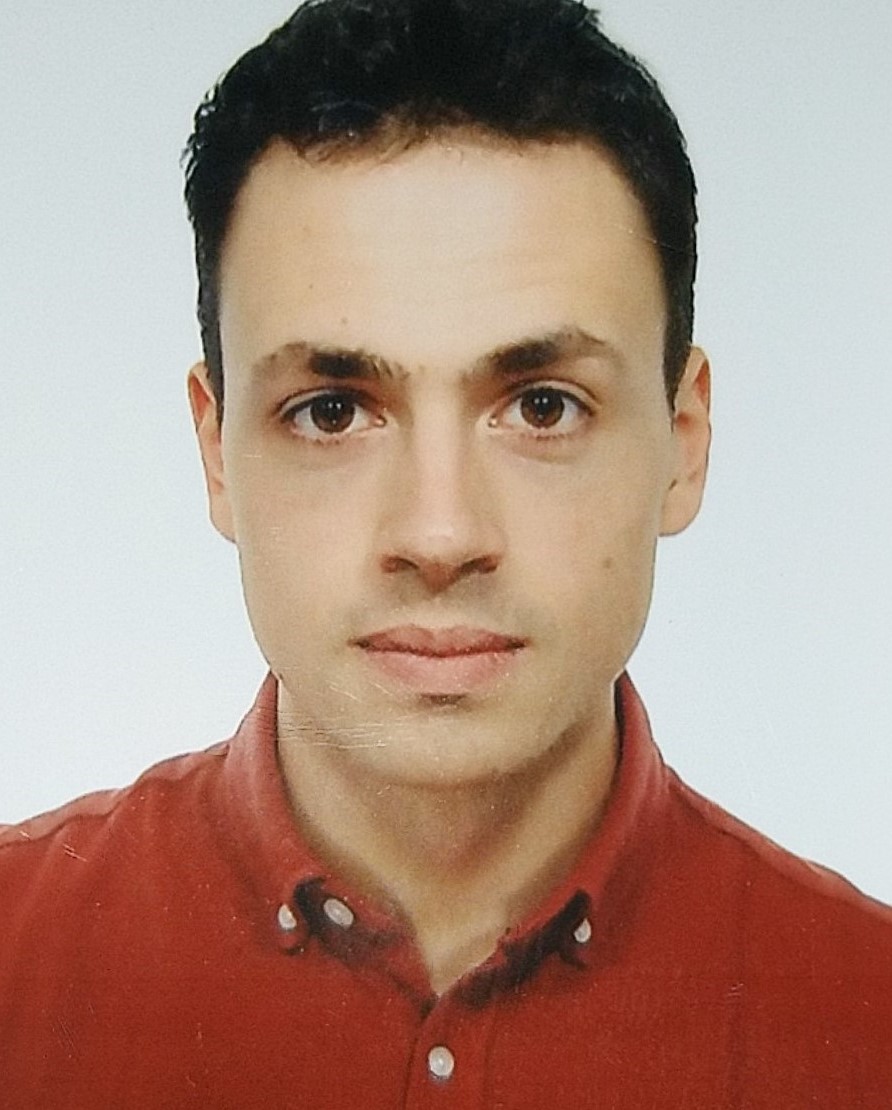}}]{Juan Ignacio Bravo Pérez-Villar} obtained a degree in Telecommunications Engineering in 2015 and received the titles belonging to the International Joint Master Program in Image Processing and Computer Vision (IPCV) in 2017 at the universities of Péter Pazmany (Hungary), Université de Bordeaux (France) and Universidad Autónoma de Madrid (Spain). Currently he is a PhD Student at the Video Processing and Understanding Lab (VPU-Lab) and a Research Engineer of the GNC/AOCS Competence Centre at DEIMOS Space. His research interests are related to visual based navigation and domain adaptation.

\end{IEEEbiography}%

\begin{IEEEbiography}[{\includegraphics[width=1in,height=1.25in,clip,keepaspectratio]{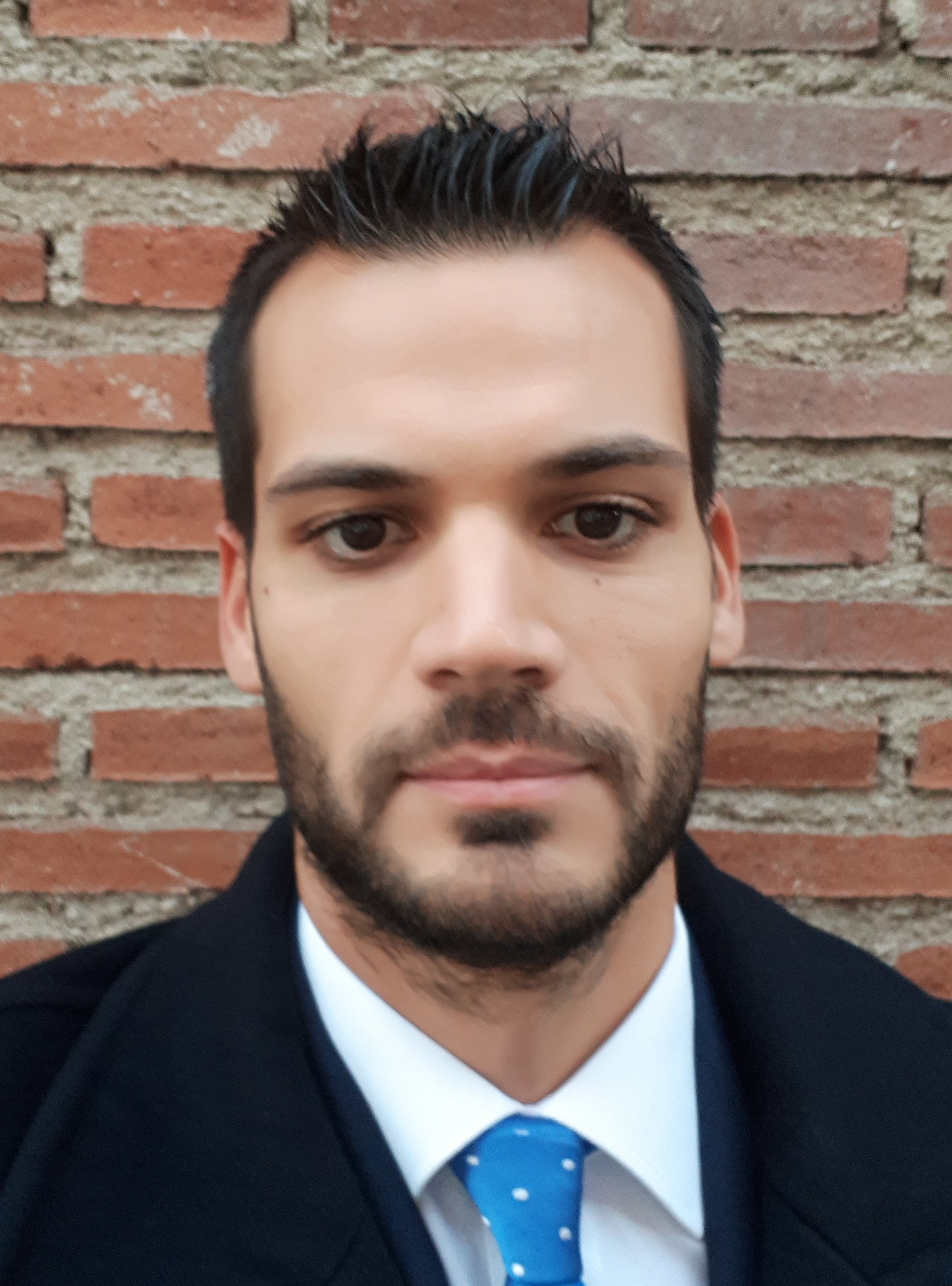}}]{Dr. Álvaro García-Martín} received the M.S. degree in Electrical Engineering ("Ingeniero de Telecomunicación" degree) in 2007 (2002-2007) and the MPhil degree in Electrical Engineering and Computer Science (postgraduate Master) in 2009 and PhD in Computer Science in 2013 at Universidad Autónoma de Madrid (Spain). From 2006 to 2019, he has been with the Video Processing and Understanding Lab (VPU-Lab) at Universidad Autónoma of Madrid as a researcher and teaching assistant. In 2008 he received a FPI research fellowship from Universidad Autónoma de Madrid. He is an Associate Professor (PhD) at Universidad Autónoma of Madrid since 2019. He has participated in several projects dealing with multimedia content transmission (PROMULTIDIS y MESH), video-surveillance (ATI@SHIVA) and activity recognition (SEMANTIC, EVENTVIDEO, HAVideo, HVD and SEGA-CV).
He also serves as a reviewer for several international Journals (IEEE TIFS, IEEE CSVT, Springer MTAP,… ) and Conferences (IEEE ICIP, IEEE AVSS,…). He has published more than 17 journal and conference papers.
His current research interests are focused in the analysis of video sequences for the video surveillance (moving object extraction, object tracking and recognition, event detection…).
\end{IEEEbiography}

\begin{IEEEbiography}[{\includegraphics[width=1in,height=1.25in,clip,keepaspectratio]{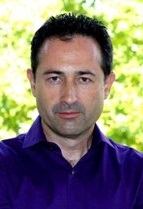}}] {Prof. Jesús Bescós} received the Ingeniero de Telecomunicación degree in 1993 and a PhD in Communications in 2001, both at Universidad Politécnica de Madrid (UPM). He was member of the Image Processing Group at UPM (1993–2002), Assistant Lecturer at this University (1997–2002), and he is now Associate Professor at Universidad Autónoma de Madrid (since 2003), where he co-leads the Video Processing and Understanding Lab. His research lines include video sequence analysis, content-based video indexing, 2D and 3D computer vision, etc. He has been actively involved in EU projects dealing with cultural heritage (e.g., RACE-Rama), education (e.g., ET-Trends), content analysis (e.g., ACTS-Hypermedia, ICTS-AceMedia and Mesh), virtual reality (e.g., IST-Slim-VRT), etc., leading to over forty publications in scientific conferences and journals. He is regular evaluator of national and international project proposals, and of submitted papers to conferences (ICIP, ICASSP, CBMI, SAMT, etc.) and journals (IEEE Tr. on CSVT, Tr. on Multimedia, Tr. on Image Processing, etc.)
\end{IEEEbiography}

\begin{IEEEbiography}[{\includegraphics[width=1in,height=1.25in,clip,keepaspectratio]{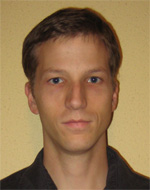}}] {Dr. Juan C. SanMiguel} received the Ph.D. degree in
computer science and telecommunication from University Autónoma of Madrid, Madrid, Spain, in 2011. He was a Post-Doctoral Researcher with Queen Mary University of London, London, U.K., from 2013 to 2014, under a Marie Curie IAPP Fellowship. He is currently Associate Professor at University Autónoma of Madrid and Researcher with the Video Processing and Understanding Laboratory. His research interests include computer vision with a focus on online performance evaluation and multicamera activity understanding for video segmentation and tracking. He has authored over 40 journal and conference papers.

\end{IEEEbiography}
\end{document}